\PassOptionsToPackage{table}{xcolor}
\documentclass[10pt,twocolumn]{article}
\ifdefined\pdfminorversion\pdfminorversion=7\fi
\usepackage[preprint]{antintlpaper}
\usepackage[utf8]{inputenc}
\usepackage{xurl}
\usepackage{latexsym,multirow,makecell,xspace,placeins}
\setcitestyle{authoryear,open={[},close={]}}

\definecolor{MethodBlue}{HTML}{EDF5FC}
\definecolor{GainBlue}{HTML}{DCECF8}
\definecolor{GroupGray}{HTML}{F2F3F5}
\newcommand{\method}{\textsc{WorldCompiler}}

\newcommand{\tabsection}[2]{\multicolumn{#1}{@{}l}{\itshape #2}\\}

\newcommand{\tvsrow}{\rowcolor{AntPaleStrong}}
\newcommand{\stitle}[1]{\vspace{1mm}\noindent{\bfseries #1}}

\AntTitle{WorldCompiler: Learn What the View Shows, Execute How the World Changes}
\AntRunningTitle{WorldCompiler}
\AntAuthors{%
  Yihao~Hu\texorpdfstring{\textsuperscript{1,2,*}}{}\and
  Yitong~Lin\texorpdfstring{\textsuperscript{1,*}}{}\and
  Pan~Wang\texorpdfstring{\textsuperscript{1,3,*}}{}\and
  Deyuan~Liu\texorpdfstring{\textsuperscript{2}}{}\and
  Hang~Wang\texorpdfstring{\textsuperscript{1}}{}\and
  Xianglong~Zeng\texorpdfstring{\textsuperscript{3}}{}\and
  Mingda~Wan\texorpdfstring{\textsuperscript{2}}{}\and
  Yongqi~Tong\texorpdfstring{\textsuperscript{1}}{}\and
  Jiang-Ming~Yang\texorpdfstring{\textsuperscript{1}}{}\and
  Wei~Wu\texorpdfstring{\textsuperscript{1,\textdagger}}{}\and
  Zhihao~Wen\texorpdfstring{\textsuperscript{1,\textdagger}}{}}
\AntAffiliations{%
  \textsuperscript{1}Ant International \quad
  \textsuperscript{2}Westlake University \quad
  \textsuperscript{3}University of Science and Technology of China}
\AntContact{\textsuperscript{*}Equal contribution. \quad
  \textsuperscript{\textdagger}Corresponding authors: Wei Wu and Zhihao Wen}
\AntDate{}
\AntAbstract{
Streaming video models answer questions while a scene unfolds, and many answers depend on what is no longer in view, such as a plant seen before the camera turned away. Answering therefore requires a world state that every observation updates, yet the same observation may increment a count, establish a lasting fact, or revise a transient attribute. Current systems keep more of the past accessible through memory, retrieval, or compression, but direct-answer supervision still makes a model decide in one step what an observation shows and how it changes that state. We call this coupling \emph{transition entanglement} and argue that it, not access to past frames, limits streaming understanding. We introduce \method{}, which separates the two: a vision-language model predicts only a typed update, and a compiler applies it to a persistent world state under explicit temporal laws. A single frozen system surpasses every published model on StreamingBench, OVO-Bench, and OST-Bench (87.01, 70.09, and 60.29), gaining 16 to 17 points on OVO-Bench exactly where answers depend on earlier observations. Learning what the view shows and executing how the world changes makes streaming state accurate, inspectable, and testable.
}

\begin{document}
\raggedbottom
\makeanttitle

\begin{figure}[!t]
\centering
\setlength{\abovecaptionskip}{4pt}
\includegraphics[width=\columnwidth]{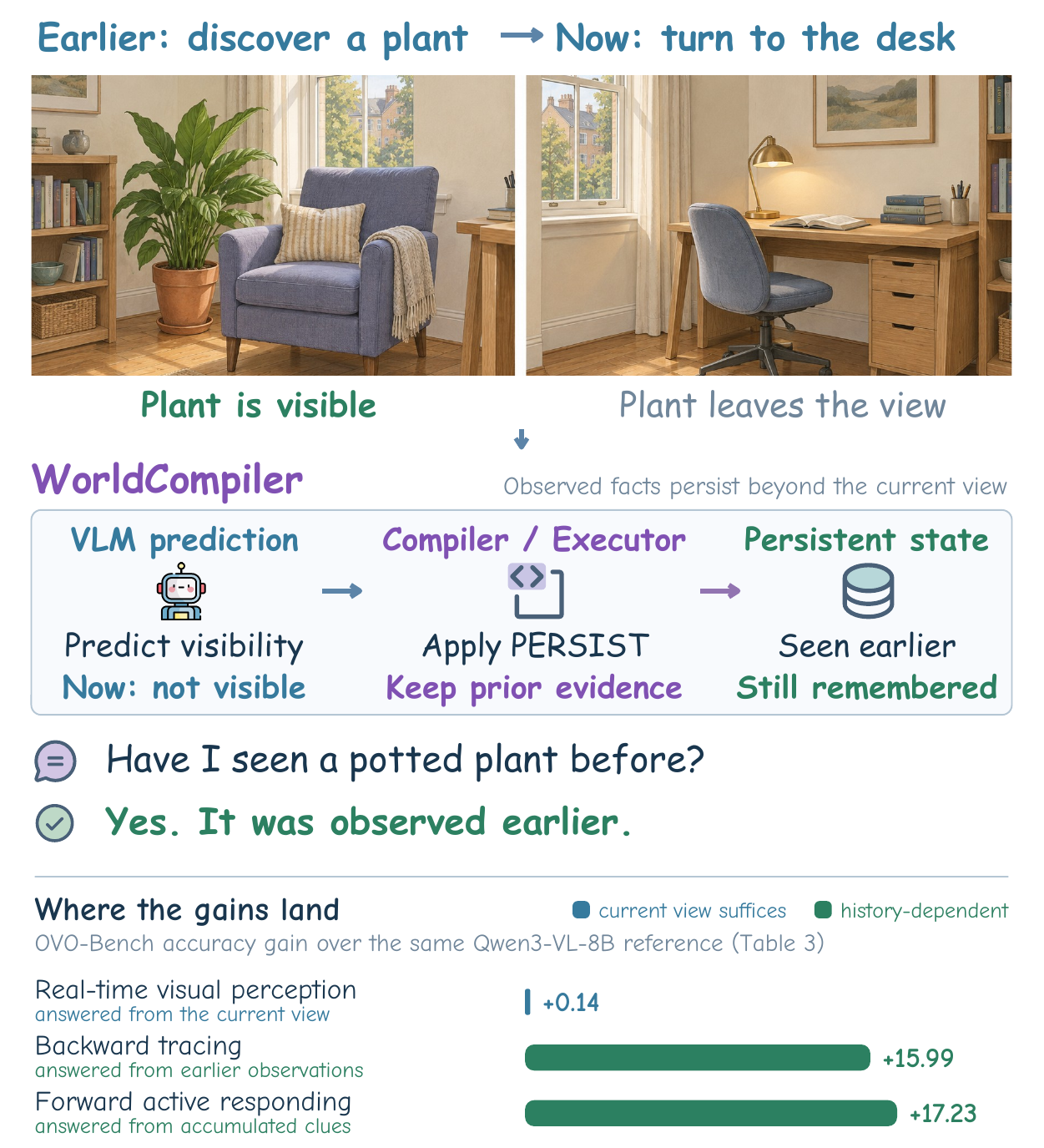}
\vspace{-2mm}
\caption{\textbf{Answer from state, not only the current view.} Illustrative persistence of an observed fact, with measured OVO-Bench gains over the matched reference (Table~\ref{tab:summary}).}
\label{fig:worldcompiler-teaser}
\vspace{-4mm}
\end{figure}

\section{Introduction}
\label{sec:intro}

Streaming video models observe a scene causally and answer questions at arbitrary timestamps \citep{lin2024streamingbench,li2025ovobench,lin2025ostbench,huang2025videochatonline,yan2025streamformer,xia2025streamo}. Consider the observer in Figure~\ref{fig:worldcompiler-teaser} (top): after passing a potted plant, the observer turns toward a desk and is asked whether a plant has been seen before. The plant is absent from the current view, yet the correct answer depends on a fact established earlier. Streaming understanding thus requires interpreting new observations while keeping established facts available to later queries.

Current streaming systems approach this problem largely through \emph{evidence access}. They retain recent frames, compress visual history, retrieve earlier observations, or maintain external memories \citep{song2024moviechat,ren2024timechat,he2024malmm,zhang2024flashvstream,huang2025videochatonline,wen2026eventmemagent,liang2026oasis,xiao2026mukv,zeng2026onestreamer}; long-video models pursue long-context scaling, adaptive compression, hierarchical memory, and agentic exploration \citep{chen2024longvila,shen2024longvu,yang2025vca,yin2026videoarm}. These mechanisms decide which evidence stays accessible but leave its effect on state implicit. Yet the same event may increment a count, establish a persistent fact, fix a first occurrence, or revise a transient attribute, and each role induces a different law of state evolution.

We call the implicit coupling between visual interpretation and history-dependent state evolution \emph{transition entanglement}. Direct-answer supervision constrains the final response, not the update that new evidence establishes. Can perceptual updates and temporal state dynamics be represented and evaluated as distinct components? Section~\ref{sec:motivation} examines history access, readout interfaces, and persistent execution under controlled settings.
Maintaining executable state from a causal stream presents three challenges. \textbf{(C1)} \emph{Perceptual grounding} requires learning typed values whose meanings agree with the supported state operations. \textbf{(C2)} \emph{Temporal execution} requires applying declared transition laws, preserving state across observations, and isolating independent sources. \textbf{(C3)} \emph{Temporal reliability} requires evaluating learned updates for correctness and consistency across temporal partitions.

These challenges arise because the required prediction depends jointly on the visual evidence and on the declared semantics of the state variable it updates: a new action is a local increment for one query and part of a cumulative estimate for another.
We introduce \method{}, which separates perception from dynamics: a vision-language model predicts typed values from causal observations, and a bounded compiler binds them to declared predicates and applies explicit temporal laws to persistent state (Figure~\ref{fig:main_framework}). In Figure~\ref{fig:worldcompiler-teaser} (middle), the model predicts only that the plant is \emph{not visible now}, and the \textsc{Persist} law keeps the earlier observation, so the state still answers \emph{yes}. One typed value serves as supervision target, execution input, and object of temporal diagnosis, so controlled interventions can examine learning and execution separately.
Our contributions are threefold:
\begin{itemize}[leftmargin=*,topsep=-1pt,itemsep=1.5pt,parsep=0pt,partopsep=0pt]
    \item We formulate \emph{transition entanglement} and a factorization of perception and dynamics that distinguishes observation-grounded predictions from their temporal state consequences (Section~\ref{sec:motivation}).
    \item We introduce \method{}, which connects typed visual supervision with executable state transitions; matched training reveals the importance of the typed output interface, and fixed-prediction interventions isolate the contribution of persistent execution (C1, C2; Section~\ref{sec:method}).
    \item We show that \method{} achieves 87.01, 70.09, and 60.29 on StreamingBench, OVO-Bench, and OST-Bench, surpassing every published model, with the largest gains on history-dependent queries (Figure~\ref{fig:worldcompiler-teaser}, bottom); temporal-partition diagnostics separate perceptual correctness, cross-partition consistency, and temporal-scale transfer (C3; Section~\ref{sec:experiments}).
\end{itemize}

\section{Diagnosing Transition Entanglement}
\label{sec:motivation}

If transition entanglement matters, historical access alone should not determine performance: the interface that turns observations into state should also contribute. Table~\ref{tab:motivation} intervenes on access, persistence, and execution semantics under controlled settings.

\begin{table}[!ht]
\centering
\footnotesize
\setlength{\tabcolsep}{5pt}
\renewcommand{\arraystretch}{1.12}
\begin{tabularx}{\columnwidth}{@{}>{\raggedright\arraybackslash}Xrr@{}}
\toprule
\textbf{Condition} & \textbf{Score (\%)} & $\boldsymbol{\Delta}$\\
\midrule

\rowcolor{AntPale}
\multicolumn{3}{@{}l@{}}{\bfseries (a) OST-Bench: evidence access vs.\ state interface}\\
\textsc{Direct} & 58.53 & --\\
\textsc{Full-History} & 57.97 & $-0.56$\\
\tvsrow
\textbf{\textsc{Typed State}} & \textbf{60.26} & $\mathbf{+1.73}$\\

\addlinespace[4pt]
\rowcolor{AntPale}
\multicolumn{3}{@{}l@{}}{\bfseries (b) OVO-Bench: identical values, local vs.\ persistent}\\
\textsc{Direct} & 64.51 & $-0.23$\\
\textsc{Local} & 64.74 & --\\
\tvsrow
\textbf{\textsc{Persistent}} & \textbf{67.06} & $\mathbf{+2.32}$\\

\addlinespace[4pt]
\rowcolor{AntPale}
\multicolumn{3}{@{}l@{}}{\bfseries (c) OVO-Bench: same transactions, varied execution}\\
\tvsrow
\textbf{\textsc{Chronological}} & \textbf{67.62} & --\\
\textsc{Reset} & 65.20 & $-2.42$\\
\textsc{Reversed} & 63.07 & $-4.55$\\
\bottomrule
\end{tabularx}
\vspace{-2mm}
\caption{\textbf{Historical access alone does not determine state evolution.} Panel (a) compares interfaces within one trained model; (b--c) intervene on fixed predictions from separate OVO models. $\Delta$ is relative to (--).}
\label{tab:motivation}
\vspace{-2mm}
\end{table}

\stitle{History access and state interpretation differ.}
Giving a trained OST model full history through a generic memory interface performs no better than direct dialogue, whereas its typed persistent interface performs better (Table~\ref{tab:motivation}a). This separates evidence availability from state interpretation.

\stitle{Persistence improves fixed predictions.}
Holding one OVO model's typed values fixed, persistent execution improves local readout by 2.32 points (Table~\ref{tab:motivation}b). The gain arises after perceptual generation, from composing the same values across time.

\stitle{Temporal semantics change the result.}
On a separate operator-balanced OVO model, reset and reversed-order replay reduce accuracy relative to chronological execution while preserving every generated transaction (Table~\ref{tab:motivation}c; Appendix~\ref{app:operator-balanced}).

\stitle{From evidence access to state execution.}
Together, these controls distinguish historical evidence access, state-readout interfaces, and temporal execution. Persistent dynamics can change the resulting answers even when perceptual predictions remain fixed.

\section{Method: \method{}}
\label{sec:method}

\begin{figure*}[t]
\centering
\setlength{\abovecaptionskip}{6pt}
\includegraphics[width=\textwidth]{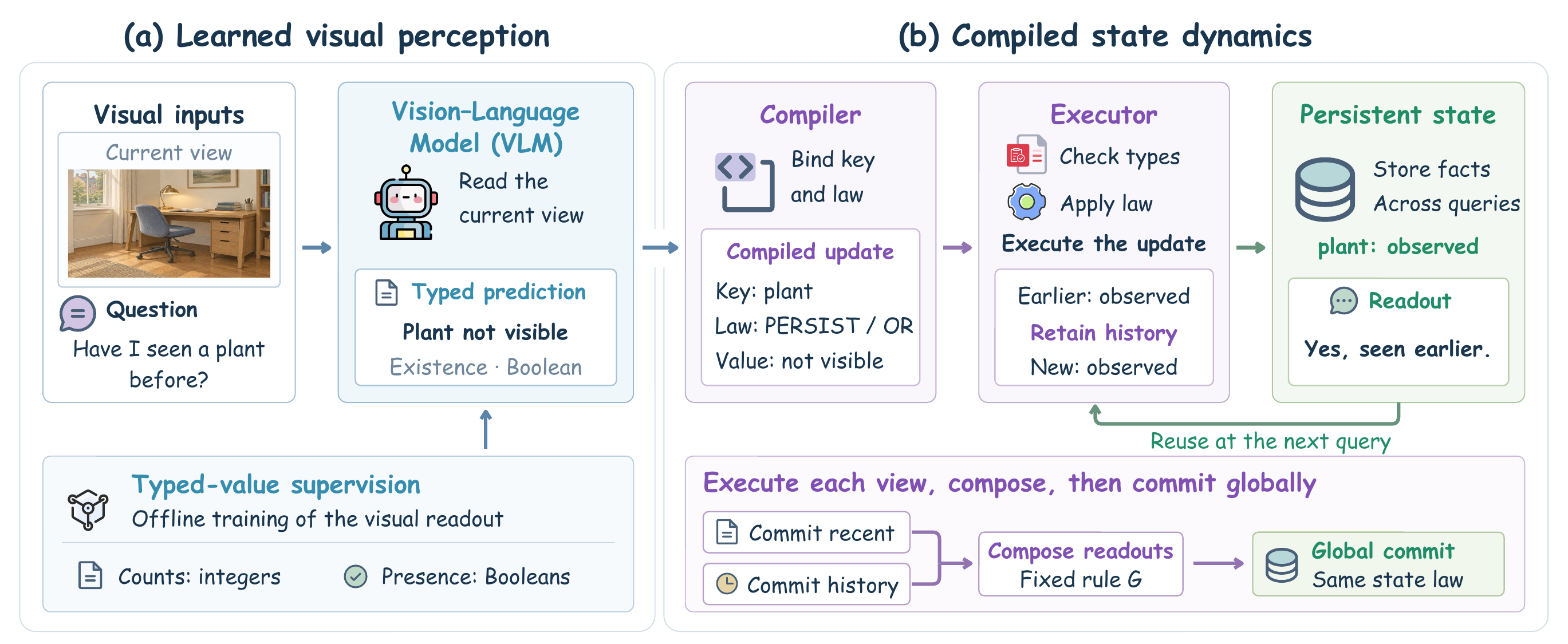}
\caption{\textbf{Learned perception and executable dynamics.} (a) The VLM learns typed values. (b) The compiler binds values to predicates and temporal laws; execution updates persistent state. View-local readouts are composed into shared global state. The Boolean example is illustrative.}
\label{fig:main_framework}
\vspace{-4pt}
\end{figure*}

\method{} addresses transition entanglement by separating
what an observation establishes from how that information
changes the world state. A learned perceptual model predicts
typed values, while a compiler executes declared temporal
laws over persistent state (Figure~\ref{fig:main_framework}).
We first formulate this separation, then introduce the
executable interface, multi-view execution, and
state-oriented supervision.

\subsection{Perception--Dynamics Factorization}
\label{sec:method:factorization}

Let $S_t$ denote the executable state summarizing
the task-relevant consequences of causal observation history.
Given new visual evidence $v_t$ and query $q_t$,
the perceptual model predicts a typed value:
\begin{equation}
    z_t = P_\theta(v_t,q_t).
    \label{eq:perception}
\end{equation}

The value $z_t$ captures what the new evidence
establishes, such as an event increment, a presence
indicator, or a temporal estimate. Its effect depends
on the declared operator $o_t$ and the previous state:
\begin{equation}
    S_{t+1} = U_{o_t}(S_t,z_t).
    \label{eq:factorization}
\end{equation}

State sufficiency is defined relative to the supported
operations. Two observation histories may share a retained
state when their differences cannot change any declared
readout. Conversely, histories requiring different updates
should remain distinguishable at the perceptual interface,
even when their current answers coincide. This shifts the
representation objective from reconstructing the complete
visual history to preserving its task-relevant consequences
under explicit temporal laws. The resulting state is therefore
an operational summary of the observed world, whose scope is
determined by the predicates and interactions it supports.

Thus, the same observation can have different state
consequences under different histories
(Figure~\ref{fig:worldcompiler-teaser}). Learning targets
the observation-conditioned update; declared dynamics
propagate its consequences. Propositions~\ref{prop:entanglement}
and~\ref{prop:state_recursion} formalize why answer supervision
may merge required updates and when correct updates preserve
a sufficient reference state.

\subsection{Typed Executable State}
\label{sec:method:interface}

Each state-bearing interaction declares a public
predicate key $k_t$, value domain $\mathcal{Z}_{k_t}$,
and temporal operator $o_t$. These declarations
are determined from the public question interface,
with source identity defining the state namespace.
Normalized questions can reuse keys across video
queries, while recognized OST questions bind
entity-level predicates.

State reuse follows predicate identity: only updates
to the same source-local key interact. This isolates
unrelated histories while allowing repeated queries to
accumulate evidence about a shared condition. The state
is therefore task-relative, following declared predicates
rather than unrestricted semantic equivalence between questions.

The VLM predicts the visually uncertain value
$z_t$. The compiler binds this value to its declared
predicate and temporal semantics:
\begin{equation}
    u_t = C(k_t,o_t,z_t).
    \label{eq:compiled-update}
\end{equation}

After validating $u_t$ against the predicate's type
and operator contract, the executor applies it:
\begin{equation}
    S_{t+1} = E(S_t,u_t).
    \label{eq:interface}
\end{equation}

The updated state then serves the task readout:
\begin{equation}
    \hat y_t = R(S_{t+1},q_t).
    \label{eq:state-readout}
\end{equation}

Accepted updates modify only the addressed source-local
variable and can be replayed deterministically; invalid
transactions preserve the previous state. The compiler owns
predicate identity, type, law, and namespace, while the VLM
supplies the observation-grounded value.

The operator determines both temporal behavior and the
required prediction: \textsc{Add} consumes a local increment,
\textsc{Max} an absolute cumulative estimate,
\textsc{Persist} an occurrence flag, \textsc{First} an earliest
timestamp, and \textsc{Replace} a revisable value. These
contracts prevent numerically similar predictions from being
treated as interchangeable evidence. Execution determines
how an accepted value changes state; whether the value is
correct remains perceptual. Appendix~\ref{app:operators}
gives the exact laws, whose compatible updates compose as
described by Proposition~\ref{prop:partition}.

A temporal law also specifies which later evidence can
revise state. Repeated cumulative estimates do not create
new increments under \textsc{Max}, while \textsc{Replace}
can revise a transient attribute. This distinction matters
when observations overlap or an object leaves view: the
same prediction must be interpreted relative to the
variable's temporal role.

Compared with unrestricted executable-program
generation, the typed interface improves interface validity:
the same fixed model produces 338 valid programs but
938 valid typed updates among 938 eligible
OVO-Bench queries
(Appendix~\ref{app:runtime-controls}).

This yields \emph{representation--execution alignment}:
the same typed value is supervised, executed, and diagnosed.

\subsection{State-Aware Multi-View Execution}
\label{sec:method:multiview}

A single causal view may expose only part of the
relevant evidence. \method{} therefore combines recent
visual detail with broader causal history. Each view
updates an independent state under the same declared
temporal semantics; their state-aware readouts are then
composed to maintain shared global state. This preserves
established facts across partial observations while
allowing complementary evidence to contribute.

Let $r_t^{(m)}$ denote the state-aware readout
obtained by executing view $m$ against its
view-local state. A fixed composition function
combines these readouts:
\begin{equation}
    \tilde z_t
    =
    G\!\left(
        \{r_t^{(m)}\}_{m=1}^{M}
    \right).
    \label{eq:view-composition}
\end{equation}

The composed proposal is interpreted according
to the applicable global operator contract.
After validation, it is committed to the
global persistent register:
\begin{equation}
    S_{t+1}
    =
    E\!\left(
        S_t,
        C(k_t,o_t,\tilde z_t)
    \right).
    \label{eq:composition}
\end{equation}

We call this \emph{state-aware composition}.
View-local execution interprets each proposal
under its temporal semantics, while global
execution maintains the composed state across
subsequent interactions.

The two-stage deployment uses \textsc{Max} and
\textsc{Persist}: their idempotent laws accommodate
cumulative state-aware values. Treating an accumulated
count as a new \textsc{Add} increment would duplicate
evidence, so local-increment \textsc{Add} is evaluated
separately in the counting diagnostics. This preserves
operator contracts across local and global execution.

The causal views and composition rule remain fixed during
evaluation; Appendix~\ref{app:frozen-bundle} reports their
implementation and computational cost.

\subsection{Learning Perceptual State Updates}
\label{sec:method:learning}

Training directly supervises the typed value
consumed by executable state transitions.
Given a target $z_t^\star$, we minimize
\begin{equation}
    \mathcal{L}_{\mathrm{state}}
    =
    -\log P_\theta(z_t^\star\mid v_t,q_t).
    \label{eq:state_loss}
\end{equation}

The target follows the declared operator contract.
For example, when prior counts of zero and two
are followed by the same newly completed action,
both local targets are $\delta=1$, while execution
produces counts of one and three. The perceptual
target represents the new increment; the executor
accounts for accumulated history.

Accordingly, interval-count tasks supervise local
increments, whereas cumulative-count queries
use prefix estimates. This distinguishes what
the model must infer visually from what the
persistent state already records.

State-bearing examples are trained with ordinary question
answering (Appendix~\ref{app:training}). Typed targets omit
the predicate/operator wrapper, averaging 17.34 rather than
66.44 characters, while matching full-transaction OST-Bench
accuracy (61.15 versus 60.87). Under a shared value-only
readout, however, the ordering reverses
(Table~\ref{tab:controlled-ablation}a); the supervision effect
therefore includes its matched readout interface.

\section{Experiments}
\label{sec:experiments}

\subsection{Experimental Setup}

We evaluate \method{} on the complete populations of StreamingBench, OVO-Bench, and OST-Bench, containing 2,495 queries, 3,035 queries, and 1,386 scenes with 9,528 scored items, respectively. We follow their causal observation protocols and official metrics using a Qwen3-VL-8B backbone.

We ask four questions: (\textbf{RQ1}) where performance improves; (\textbf{RQ2}) how supervision and readout jointly affect state-oriented prediction (C1); (\textbf{RQ3}) what persistent execution contributes under fixed perceptual predictions (C2); and (\textbf{RQ4}) whether learned updates remain correct and stable across temporal partitions (C3). Appendix~\ref{app:experimental-protocol} provides training and evaluation details.

\begin{table*}[!t]
\centering
\scriptsize
\setlength{\tabcolsep}{7pt}
\renewcommand{\arraystretch}{1.0}
\begin{tabular}{@{}lc *{10}{c} c@{}}
\toprule
Model & Frames
  & OP & CR & CS & ATP & EU & TR & PR & SU & ACP & CT
  & ALL \\
\midrule
Human & --
  & 89.47 & 92.00 & 93.60 & 91.47 & 95.65 & 92.52 & 88.00 & 88.75 & 89.74 & 91.30
  & 91.46 \\
\midrule
\tabsection{13}{Proprietary MLLMs}
Gemini~1.5 Pro & 1\,fps
  & 79.02 & 80.47 & 83.54 & 79.67 & 80.00 & 84.74 & 77.78 & 64.23 & 71.95 & 48.70
  & 75.69 \\
GPT-4o & 64
  & 77.11 & 80.47 & 83.91 & 76.47 & 70.19 & 83.80 & 66.67 & 62.19 & 69.12 & 49.22
  & 73.28 \\
Claude~3.5 Sonnet & 20
  & 80.49 & 77.34 & 82.02 & 81.73 & 72.33 & 75.39 & 61.11 & 61.79 & 69.32 & 43.09
  & 72.44 \\
\midrule
\tabsection{13}{Open-source offline MLLMs}
InternVL-V2-8B & 16
  & 68.12 & 60.94 & 69.40 & 77.12 & 67.70 & 62.93 & 59.26 & 53.25 & 54.96 & 56.48
  & 63.72 \\
Qwen2-VL-7B & 0.2--1\,fps
  & 75.20 & 77.45 & 73.19 & 82.81 & 68.32 & 71.03 & 72.22 & 61.19 & 61.47 & 46.11
  & 69.04 \\
Qwen3-VL-8B & 0.2--1\,fps
  & 73.57 & 71.09 & 76.34 & 80.39 & 65.22 & 74.77 & 78.70 & 67.89 & 61.47 & 47.67
  & 70.20 \\
\midrule
\tabsection{13}{Open-source online MLLMs}
Flash-VStream-7B & --
  & 25.89 & 43.57 & 24.91 & 23.87 & 27.33 & 13.08 & 18.52 & 25.20 & 23.87 & 48.70
  & 23.23 \\
VideoLLM-online-8B & 2\,fps
  & 39.07 & 40.06 & 34.49 & 31.05 & 45.96 & 32.40 & 31.48 & 34.16 & 42.49 & 27.89
  & 35.99 \\
Dispider-7B & 1\,fps
  & 74.92 & 75.53 & 74.10 & 73.08 & 74.44 & 59.92 & 76.14 & 62.91 & 62.16 & 45.80
  & 67.63 \\
StreamForest-7B & 1\,fps
  & 83.11 & 82.81 & 82.65 & 84.26 & 77.50 & 78.19 & 76.85 & 69.11 & 75.64 & 54.40
  & 77.26 \\
Mage-VL-4B$^{\ast}$ & --
  & 88.3 & 71.9 & 91.2 & 89.8 & 75.6 & 90.7 & 83.3 & 81.7 & 81.9 & 40.9
  & 82.1 \\
JoyAI-VL-Interaction-8B$^{\ast}$ & --
  & 87.2 & 83.6 & 92.1 & 88.9 & 81.9 & 91.3 & 88.0 & 74.0 & 81.3 & 46.1
  & 82.7 \\
VideoChat3-4B$^{\ast}$ & --
  & 86.4 & 84.4 & 93.7 & 88.5 & 81.3 & 83.8 & 78.7 & 78.9 & 83.6 & 42.0
  & 81.9 \\
\midrule
\tabsection{13}{Online video agents}
StreamAgent-7B & 1\,fps
  & 79.63 & 78.31 & 79.28 & 75.87 & 74.74 & 76.92 & 82.94 & 66.31 & 73.69 & 55.40
  & 74.28 \\
EventMemAgent & $\le$32
  & 83.92 & 76.56 & 87.70 & 85.29 & 77.02 & 79.44 & 77.78 & 68.29 & 72.24 & 48.70
  & 77.00 \\
\midrule
\tabsection{13}{Ours}
Qwen3-VL-8B-Instruct & 4
  & 84.74 & 71.88 & 90.85 & 89.44
  & 76.10 & 89.10 & 81.48 & 78.05
  & 79.04 & 41.97
  & 80.52 \\
\tvsrow
\textbf{\method{}} & $\boldsymbol{\leq 32}$
  & 90.74 & 87.50 & 94.64
  & 85.84 & 83.02 & 95.33
  & 87.04 & 80.89 & 93.73
  & 55.96
  & 87.01 \\
\bottomrule
\end{tabular}
\caption{Locked-population StreamingBench accuracy by official task. Frames is the per-call maximum; published rows retain released protocols. Our population has 2,495 queries. $^{\ast}$\citet{zeng2026onestreamer}, Table~9 (2,498 queries); --: unreported.}
\label{tab:sb-compare}
\end{table*}

\subsection{Overall Performance (RQ1)}
\label{sec:exp-overall}

\stitle{Strong performance across three streaming benchmarks.}
The complete system reaches 87.01 on StreamingBench, 70.09 on OVO-Bench, and 60.29 on OST-Bench (Tables~\ref{tab:sb-compare}, \ref{tab:summary}, and~\ref{tab:ovo-compare}), improving the matched Qwen3-VL-8B-Instruct reference by 6.49, 11.12, and 9.77 points. Published rows provide context under their released protocols; the controlled comparisons below provide attribution.

\stitle{Gains concentrate on history-dependent tasks.}
On OVO-Bench, backward tracing and forward active responding improve by 15.99 and 17.23 points over the matched reference, while real-time visual perception changes by only 0.14 points (Table~\ref{tab:summary}). The strongest gains occur when answers require previously established conditions or accumulated clues, rather than recognition of the current frame alone.

A similar pattern appears on StreamingBench, where nine of ten task families improve. Causal reasoning and action perception gain 15.62 and 14.69 points, respectively, whereas attribute perception decreases by 3.60 points. OST-Bench improves in fourteen of fifteen subtypes, with a 22.49-point gain in direction estimation. These patterns suggest that the principal benefit lies in maintaining information across observations; the controls below separate this effect from the complete system.

\begin{table}[!t]
\centering
\scriptsize
\setlength{\tabcolsep}{2pt}
\renewcommand{\arraystretch}{1.12}
\begin{tabularx}{\columnwidth}{@{}>{\raggedright\arraybackslash}Xrrrrr@{}}
\toprule
& \multicolumn{4}{c}{OVO-Bench} & OST-Bench\\
\cmidrule(lr){2-5}\cmidrule(lr){6-6}
Model & RT & BT & FAR & Overall & Avg\\
\midrule
Selected published & \textbf{81.7} & 62.54 & 58.3 & 64.02 & 54.2\\
Reference & 81.38 & 51.45 & 44.09 & 58.97 & 50.52\\
\tvsrow
\textbf{\method{}} & 81.52 & \textbf{67.44} & \textbf{61.32} & \textbf{70.09} & \textbf{60.29}\\
\midrule
$\Delta$ vs.\ reference & $+0.14$ & $+15.99$ & $+17.23$ & $+11.12$ & $+9.77$\\
$\Delta$ vs.\ selected published & $-0.18$ & $+4.90$ & $+3.02$ & $+6.07$ & $+6.09$\\
\bottomrule
\end{tabularx}
\caption{\textbf{Where the gains land.} Official OVO groups and OST score. Reference uses our protocol; ``Selected published'' is the strongest displayed contextual result per column under released protocols (Table~\ref{tab:ovo-compare}; Appendix~\ref{app:per-task}).}
\label{tab:summary}
\end{table}

\begin{table*}[!t]
\centering
\scriptsize
\setlength{\tabcolsep}{1.3pt}
\renewcommand{\arraystretch}{1.0}
\resizebox{\linewidth}{!}{%
\begin{tabular}{@{}lc *{7}{c} *{4}{c} *{4}{c} c@{}}
\toprule
& & \multicolumn{7}{c}{Real-Time Visual Perception}
  & \multicolumn{4}{c}{Backward Tracing}
  & \multicolumn{4}{c}{Forward Active Responding} & \\
\cmidrule(lr){3-9}\cmidrule(lr){10-13}\cmidrule(lr){14-17}
Model & \makecell{Frames\\per call}
  & OCR & ACR & ATR & STU & FPD & OJR & Avg.
  & EPM & ASI & HLD & Avg.
  & REC & SSR & CRR & Avg.
  & Overall \\
\midrule
Human & --
  & 93.96 & 92.57 & 94.83 & 92.70 & 91.09 & 94.02 & 93.20
  & 92.59 & 93.02 & 91.37 & 92.33
  & 95.48 & 89.67 & 93.56 & 92.90
  & 92.81 \\
\midrule
\tabsection{18}{Proprietary MLLMs}
Gemini~3 Flash & --
  & 86.58 & 68.81 & 79.31 & 58.99 & 70.30 & 70.65 & 72.44
  & 58.92 & 76.35 & 52.15 & 62.47
  & 35.53 & 74.24 & 61.67 & 57.15
  & 64.02 \\
Gemini~1.5 Pro & 1\,fps
  & 85.91 & 66.97 & 79.31 & 58.43 & 63.37 & 61.96 & 69.32
  & 58.59 & 76.35 & 52.64 & 62.54
  & 35.53 & 74.24 & 61.67 & 57.15
  & 63.00 \\
GPT-4o & 64
  & 69.80 & 64.22 & 71.55 & 51.12 & 70.30 & 59.78 & 64.46
  & 57.91 & 75.68 & 48.66 & 60.75
  & 27.58 & 73.21 & 59.40 & 53.40
  & 59.54 \\
\midrule
\tabsection{18}{Open-source offline MLLMs}
Qwen2-VL-7B & 64
  & 60.40 & 50.46 & 56.03 & 47.19 & 66.34 & 55.43 & 55.98
  & 47.81 & 35.48 & 56.08 & 46.46
  & 31.66 & 65.82 & 48.75 & 48.74
  & 50.39 \\
InternVL-V2-8B & 64
  & 67.11 & 60.55 & 63.79 & 46.07 & 68.32 & 56.52 & 60.39
  & 48.15 & 57.43 & 24.73 & 43.44
  & 26.50 & 59.14 & 54.14 & 46.60
  & 50.15 \\
Qwen3-VL-8B & 64
  & 75.17 & 58.72 & 72.41 & 57.30 & 70.30 & 59.24 & 65.52
  & 56.57 & 69.59 & 12.37 & 46.18
  & 47.13 & 67.57 & 52.50 & 55.73
  & 55.81 \\
\midrule
\tabsection{18}{Open-source online MLLMs}
Flash-VStream-7B & 1\,fps
  & 24.16 & 29.36 & 28.45 & 33.71 & 25.74 & 28.80 & 28.37
  & 39.06 & 37.16 & 5.91 & 27.38
  & 8.02 & 67.25 & 60.00 & 45.09
  & 33.61 \\
Dispider-7B & 1\,fps
  & 57.72 & 49.54 & 62.07 & 44.94 & 61.39 & 51.63 & 54.55
  & 48.48 & 55.41 & 4.30 & 36.06
  & 18.05 & 37.36 & 48.75 & 34.72
  & 41.78 \\
StreamForest-7B & 1\,fps
  & 68.46 & 53.21 & 71.55 & 47.75 & 65.35 & 60.87 & 61.20
  & 58.92 & 64.86 & 32.26 & 52.02
  & 32.81 & 70.59 & 57.08 & 53.49
  & 55.57 \\
Mage-VL-4B & --
  & 96.0 & 81.7 & 86.2 & 71.3 & 72.3 & 82.6 & 81.7
  & 53.2 & 54.7 & 38.7 & 48.9
  & 19.5 & 73.3 & 41.7 & 44.8
  & 58.5 \\
JoyAI-VL-Interaction-8B & --
  & 80.5 & 60.6 & 77.6 & 65.2 & 79.2 & 62.0 & 70.9
  & 58.9 & 62.8 & 23.1 & 48.3
  & 45.8 & 70.3 & 58.8 & 58.3
  & 59.1 \\
VideoChat3-4B & --
  & 85.9 & 71.6 & 79.3 & 55.1 & 73.3 & 71.2 & 72.7
  & 56.9 & 58.1 & 47.3 & 54.1
  & 23.8 & 67.2 & 55.0 & 48.7
  & 58.5 \\
\midrule
\tabsection{18}{Online video agents}
StreamAgent-7B & 1\,fps
  & 71.20 & 53.20 & 63.60 & 53.90 & 67.30 & 58.70 & 61.30
  & 54.80 & 58.10 & 25.80 & 41.70
  & 35.90 & 48.40 & 52.00 & 45.40
  & 49.40 \\
EventMemAgent & $\le$32
  & 75.84 & 69.72 & 73.28 & 55.62 & 67.33 & 67.93 & 68.29
  & 59.60 & 70.95 & 43.55 & 58.03
  & 33.67 & 72.02 & 62.08 & 55.92
  & 60.75 \\
\midrule
\tabsection{18}{Ours}
Qwen3-VL-8B-Instruct & 4
  & 95.30 & 80.73 & 82.76
  & 73.60 & 73.27 & 82.61
  & 81.38
  & 54.55 & 62.16 & 37.63
  & 51.45
  & 19.91 & 66.93 & 45.42
  & 44.09
  & 58.97 \\
\tvsrow
\textbf{\method{}} & $\boldsymbol{\leq 32}$
  & 97.32 & 81.65 & 81.90
  & 70.22 & 79.21 & 78.80
  & 81.52
  & 67.68 & 66.89 & 67.74
  & 67.44
  & 35.39 & 74.40 & 74.17
  & 61.32
  & 70.09 \\
\bottomrule
\end{tabular}}
\caption{\textbf{OVO-Bench task-level performance (\%).}
Overall follows the official task-then-group macro.
Published results follow their respective evaluation protocols.}
\label{tab:ovo-compare}
\end{table*}

\subsection{Controlled Ablations (RQ2--RQ3)}
\label{sec:exp-ablations}
\begin{table*}[!t]
\centering
\footnotesize
\setlength{\tabcolsep}{2pt}
\setlength{\fboxsep}{0pt}
\renewcommand{\arraystretch}{1.08}

% (a) State-oriented learning
\begin{minipage}[t]{0.32\textwidth}
\centering
\colorbox{AntPale}{\parbox[c][1.65\baselineskip][c]{\linewidth}{
\centering\textbf{(a) OST: State Learning}
}}
\begin{tabularx}{\linewidth}{@{}>{\raggedright\arraybackslash}Xrr@{}}
\toprule
\textbf{Target} & \textbf{Native} & \textbf{Unified}\\
\midrule
Direct (D) & $53.46\pm1.21$ & \textbf{51.83}\\
Full txn. (F) & $60.87\pm0.71$ & 43.41\\
\tvsrow
\textbf{Typed (T)} & $\mathbf{61.15}\pm0.73$ & 47.03\\
$\Delta$ F $-$ D & $+7.41$ & $-8.42$\\
$\Delta$ T $-$ D & $\mathbf{+7.69}$ & $-4.80$\\
$\Delta$ T $-$ F & $+0.28$ & $+3.62$\\
\bottomrule
\end{tabularx}
\end{minipage}
\hfill
% (b) Final-stage execution semantics on fixed predictions
\begin{minipage}[t]{0.32\textwidth}
\centering
\setlength{\tabcolsep}{1pt}
\colorbox{AntPale}{\parbox[c][1.65\baselineskip][c]{\linewidth}{
\centering\textbf{(b) Persistent-State Semantics}
}}
\begin{tabularx}{\linewidth}{@{}>{\raggedright\arraybackslash}Xrrr@{}}
\toprule
\textbf{Condition} & \textbf{SB} & \textbf{OVO} & \textbf{OST}\\
\midrule
No global commit & 86.89 & 69.93 & 60.22\\
Reset & 86.89 & 69.93 & 60.22\\
Reversed & 86.45 & 66.40 & 60.24\\
Merged sources & 86.37 & 69.49 & 54.88\\
\tvsrow
\textbf{Chronological} & \textbf{87.01} & \textbf{70.09} & \textbf{60.29}\\
$\Delta$ vs.\ no commit & $+0.12$ & $+0.16$ & $+0.07$\\
\bottomrule
\end{tabularx}
\end{minipage}
\hfill
% (c) Multi-view execution
\begin{minipage}[t]{0.32\textwidth}
\centering
\colorbox{AntPale}{\parbox[c][1.65\baselineskip][c]{\linewidth}{
\centering\textbf{(c) OVO: Views and Execution}
}}
\begin{tabularx}{\linewidth}{@{}>{\raggedright\arraybackslash}Xr@{}}
\toprule
\textbf{Condition} & \textbf{Score (\%)}\\
\midrule
Broader-history view & 69.90\\
Recent-detail view & 69.53\\
No state & 68.45\\
Global only & 70.00\\
View-local only & 69.93\\
\tvsrow
\textbf{Both stages} & \textbf{70.09}\\
\bottomrule
\end{tabularx}
\end{minipage}

\caption{\textbf{Controlled ablations.} (a) Native: three-seed mean $\pm$ SD; unified: seed 59. (b) Fixed predictions; upstream view-local execution and composition remain enabled. Only the final global commit varies (OST: entity reconciliation). (c) OVO execution stages. $\Delta$: displayed scores (pp); 1.65 uses unrounded scores.}
\label{tab:controlled-ablation}
\end{table*}

\begin{table*}[!t]
\centering
\footnotesize
\setlength{\tabcolsep}{3pt}
\renewcommand{\arraystretch}{1.16}
\begin{tabularx}{\textwidth}{@{}>{\raggedright\arraybackslash}p{0.26\textwidth}*{7}{>{\raggedleft\arraybackslash}X}@{}}
\toprule
\rowcolor{AntPale}
& \multicolumn{4}{c}{\textbf{Source-held videos} ($N{=}63$)} & \multicolumn{3}{c}{\textbf{Full population} ($N{=}4{,}470$)}\\
\cmidrule(lr){2-5}\cmidrule(lr){6-8}
\textbf{Count pilot} & \makecell[r]{All correct\\2/4/8\,s $\uparrow$} & \makecell[r]{Pairwise\\(/189) $\uparrow$} & \makecell[r]{Mean\\range $\downarrow$} & \makecell[r]{12\,s\\correct $\uparrow$} & \makecell[r]{4\,s correct\\(\%) $\uparrow$} & \makecell[r]{All correct\\2/4/8\,s $\uparrow$} & \makecell[r]{Pairwise\\(\%) $\uparrow$}\\
\midrule
Independent multiscale & 6 & 60 & 1.937 & \cellcolor{AntPaleStrong}\textbf{28} & 33.58 & 490 & 35.92\\
Adjacent families & 6 & \cellcolor{AntPaleStrong}\textbf{69} & \cellcolor{AntPaleStrong}\textbf{1.492} & 25 & 37.32 & \cellcolor{AntPaleStrong}\textbf{575} & \cellcolor{AntPaleStrong}\textbf{42.49}\\
Adjacent, matched targets & \cellcolor{AntPaleStrong}\textbf{8} & \cellcolor{AntPaleStrong}\textbf{69} & 1.873 & 19 & \cellcolor{AntPaleStrong}\textbf{39.80} & 534 & 36.91\\
\bottomrule
\end{tabularx}
\caption{\textbf{Temporal compositionality (RQ4).} Correctness, agreement, and range across 2/4/8\,s partitions, with 12\,s transfer. Blue marks the best value per column.}
\vspace{-2mm}
\label{tab:temporal}
\end{table*}

Table~\ref{tab:controlled-ablation} examines the respective contributions of state-oriented learning, persistent execution, and multi-view state composition.

\stitle{Supervision and readout jointly shape performance (RQ2).}
\label{sec:exp-learning}
Three paired OST runs use matched data, initialization, and optimization while varying the supervision target and its corresponding readout. Typed-value training improves native accuracy from $53.46\pm1.21$ to $61.15\pm0.73$, with positive differences across all three seeds (Table~\ref{tab:controlled-ablation}a).

A unified value-only evaluation reveals a different ordering. Under a shared readout, the direct-answer, full-transaction, and typed-value models achieve 51.83\%, 43.41\%, and 47.03\%, respectively. This diagnostic uses one paired seed (59), whereas native results average three seeds.

The contrast distinguishes native effectiveness from cross-interface value recovery: state-oriented training performs best with its matched readout, while direct-answer training leads under the common readout. The measured learning effect thus includes readout matching (Appendix~\ref{app:learning-controls}); RQ3 examines execution with predictions fixed.

\stitle{Persistent dynamics independently improve fixed predictions (RQ3).}
\label{sec:exp-execution}
Holding generated typed values fixed, persistent execution improves OVO-Bench from 64.74 to 67.06, a gain of 2.32 points (95\% paired-bootstrap CI $[1.33,3.19]$; Table~\ref{tab:motivation}b). The improvement arises entirely from how the same predictions are maintained and read across time.

On the complete system, execution interventions further reveal the importance of temporal order and source identity. Resetting state recovers local-readout behavior, while reversing execution order or merging source namespaces changes the resulting answers (Table~\ref{tab:controlled-ablation}b). These effects matter when successive interactions revisit shared state variables.

The attribution experiment in Table~\ref{tab:motivation}(b) includes 938 state-bearing queries and 130 source-local keys. At the first visit, local and persistent readouts agree, each answering 100/130 queries correctly. At the third through fifth visits, persistent execution answers 143/359 correctly versus 78/359 locally. At visits six and later, the corresponding counts are 14/319 and 4/319. These are query-level accuracies within visit groups, distinct from the official task-then-group macro. The pattern connects execution gains to repeated access to shared variables, while showing that longer histories alone do not ensure larger gains.

\stitle{The benefit of typed updates is operator-dependent.}
A complementary control applies deterministic retention laws to canonical answers. Under the tested recent-view protocol, ordinary answers with \textsc{Max}/\textsc{Persist} reach 67.25 on OVO-Bench, comparable to 67.06 for typed persistent execution. Thus, monotone retention can benefit from either prediction interface.

The distinction becomes more apparent under other temporal semantics. In the operator-matched OST comparison, typed predictions outperform canonical answers by 5.58 points on \textsc{First} and 1.06 on \textsc{Replace} (Appendix~\ref{app:operator-comparison}). These descriptive results suggest that structured updates are useful when an ordinary answer does not directly express the required temporal value.

\stitle{State execution also helps across views.}
Within the same multi-view setting, complete execution improves accuracy from 68.45 to 70.09, a 1.65-point gain computed from unrounded scores (95\% paired-bootstrap CI $[0.76,2.31]$; Table~\ref{tab:controlled-ablation}c). Both single-stage variants improve over composition without state, with small additional gains from combining the stages. This separates the benefit of state execution from the choice of execution stage.

\subsection{Temporal Compositionality (RQ4)}
\label{sec:exp-composition}

Executable updates provide a structural criterion beyond answer accuracy. For adjacent intervals $I_L$ and $I_R$, a composable counting update should satisfy
\begin{equation}
    \delta(I_L\cup I_R)
    =
    \delta(I_L)+\delta(I_R).
    \label{eq:composition-law}
\end{equation}
This law concerns compatible local updates. Our experiments measure final-count behavior across temporal partitions, including the effects of visual sampling and perceptual prediction. All-partition correctness requires accurate counts at every width; pairwise agreement measures equality across partitions; prediction range measures count variation.

\stitle{Consistency and correctness favor different training choices.}
On 63 source-held videos, target-matched adjacent training improves all-partition correctness from 6/63 to 8/63 and pairwise agreement from 60/189 to 69/189 relative to independent multiscale training (Table~\ref{tab:temporal}). Mean prediction range also decreases from 1.937 to 1.873 across the 2-, 4-, and 8-second partitions.

The broader evaluation over 4,470 sources reveals a clearer distinction. Target matching improves four-second correctness from 33.58\% to 39.80\%, whereas unmatched adjacent training achieves the highest pairwise agreement (42.49\%) and all-partition correctness (575/4,470). The best training choice for single-width accuracy therefore differs from the best for cross-partition stability.

\stitle{Temporal-scale transfer remains distinct.}
At the unseen 12-second width, the ordering reverses: target-matched training produces 19/63 correct counts, compared with 28/63 for independent training. Changing window width also changes visual sampling and model calls, so these observations characterize the complete perception--execution process.

These experiments separate single-scale accuracy, cross-partition stability, and temporal-scale transfer (Appendix~\ref{app:composition}).

\begin{figure}[t]
\centering
\setlength{\abovecaptionskip}{4pt}
\includegraphics[width=\columnwidth]{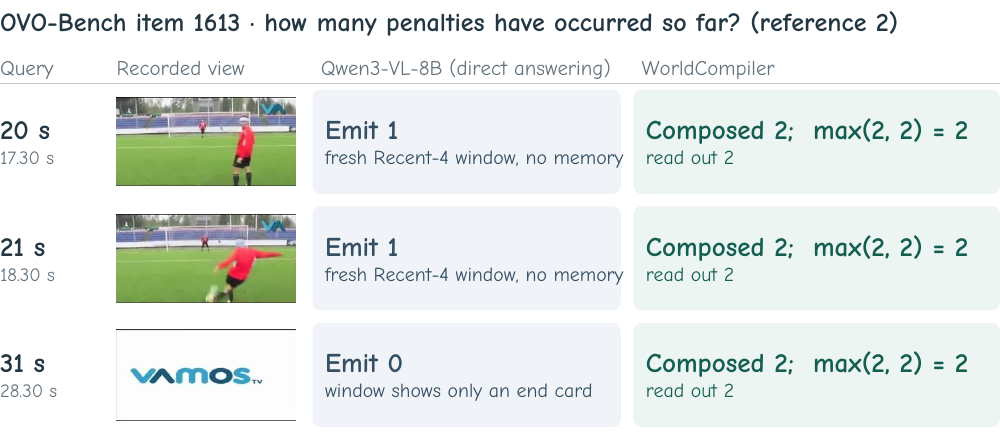}
\vspace{-2mm}
\caption{\textbf{A recorded OVO-Bench case.} Fresh-window answers are 1, 1, 0; \method{} composes 2 before \textsc{Max} retains it. Full trace: Figure~\ref{fig:case-ovo-count}.}
\vspace{-4mm}
\label{fig:case-main}
\end{figure}

\subsection{Case Study: State Across Changing Views}
\label{sec:exp-case}

In Figure~\ref{fig:case-main}, a direct reader using fresh recent windows predicts 1, 1, and 0. The complete system combines historical evidence and returns 2 even when the recent window contains only an end card. Every displayed composed proposal is already 2 before \textsc{Max} is applied.

The trace also reveals a limitation of monotone state retention. At the earlier 15-second query, the stored count was already two against a reference of one. Later events made this value correct without repairing its earlier error. Correctness at later queries therefore does not imply correctness throughout the state trajectory.

This complete-system example complements the fixed-prediction intervention in Table~\ref{tab:motivation}(b), which isolates persistence. Appendix~\ref{app:qualitative-replay} provides additional trajectories.

\section{Related Work}
\label{sec:related}

\paragraph{World models and state estimation.}
World models represent environments through learned states and dynamics for prediction, planning, and control \citep{ha2018worldmodels,hafner2019planet,hafner2020dreamer,hafner2023dreamerv3,hafner2025mastering,bruce2024genie,sun2024latent}. \method{} studies observation-conditioned maintenance of explicit world variables whose semantics follow supported predicates and declared temporal laws.

\paragraph{Streaming video memory.}
Streaming and long-video systems preserve historical evidence through compression, retrieval, recurrent representations, and adaptive access \citep{song2024moviechat,song2024moviechatplus,he2024malmm,zhang2024flashvstream,huang2025videochatonline,wang2025videollamb,liang2026oasis,xiao2026mukv,wen2026eventmemagent}. SimpleStream \citep{shen2026simplestream} and OneStreamer \citep{zeng2026onestreamer} study streaming evidence and memory. ViSS separates visible-state supervision from causal view selection \citep{hu2026viss}; ChronoReplay organizes spatial supervision as ordered episodes \citep{hu2026chronoreplay}. \method{} examines how inferred values update state, using fixed predictions to isolate execution.

\paragraph{Structured and executable reasoning.}
Dialogue state tracking maintains explicit slots and selective updates
\citep{wu2019trade,kim2020somdst,ouyang2020dstsc,heck2020trippy,le2022multimodaldst}.
Programmatic visual reasoning executes operations over perceptual outputs
\citep{andreas2016nmn,yi2018nsvqa,gupta2023visprog,suris2023vipergpt,subramanian2023codevqa}.
Trajectory2Gym grounds tool-use training actions in agent-visible histories
\citep{hu2026trajectory2gym}. \method{} learns typed visual updates for
persistent temporal execution (Appendix~\ref{app:related}).

\section{Conclusion}
\label{sec:conclusion}

We introduced \method{}, which separates learned perceptual updates from executable temporal dynamics for streaming video understanding. Matched training reveals how supervision and readout jointly shape native performance, while fixed-prediction experiments establish an independent benefit from persistent execution. Operator-matched comparisons identify where structured updates offer value beyond answer accumulation, and temporal-partition diagnostics distinguish compositional consistency from perceptual correctness. Together, these findings clarify the complementary roles of learning what observations establish and executing how those facts evolve into persistent world state.

\section*{Limitations}
\method{}'s state semantics are declared rather than learned: predicate keys and temporal operators come from deterministic rules over public question wording, and the operator library covers five temporal roles (\textsc{Add}, \textsc{Max}, \textsc{Persist}, \textsc{First}, \textsc{Replace}). This keeps the compiler free of task labels and makes every state trajectory replayable, but it restricts the interface to questions whose state semantics such rules recognize. A natural next step is to let a small declaration model propose the key, type, and operator for an open-ended question and have the compiler validate the proposal against the typed state, so that the learned part grows while execution stays deterministic; new operators can join the same algebra as long as they remain associative (Proposition~\ref{prop:partition}).

Second, learned updates do not yet transfer across temporal scales: adjacent-family training improves correctness and consistency at 2--8-second partitions, but the ordering reverses at 12 seconds (Section~\ref{sec:exp-composition}), and counting remains the weakest StreamingBench category. Because the executable interface exposes the composition law directly, two remedies are concrete: train over a wider, randomized range of window widths with composition-matched targets, and turn Eq.~\ref{eq:composition-law} into a consistency loss that penalizes disagreement between an interval's predicted update and the sum over its sub-intervals, so that scale-stable updates are learned rather than assumed.

\section*{Ethics Statement}
Persistent spatial memory may retain sensitive information about people and environments. Deployments should minimize stored observations, restrict state fields to the task, retain provenance, provide access and deletion controls, and avoid using inferred personal identities as entity keys. The primary typed interface executes declared state operators rather than arbitrary model-generated code.

The training sources and benchmark protocols described in this paper are used for research evaluation. Training and evaluation follow the source and split assignments documented in Appendices~\ref{app:training} and~\ref{app:experimental-protocol}.

\bibliographystyle{plainnat}
\bibliography{references}

\appendix
\section{Executable State Interface and Deployment Protocol}
\label{app:state-interface}

This section specifies the concrete executable-state interface used by \method{} and the frozen deployment protocol underlying the complete benchmark evaluation. The main text treats state transitions abstractly; here we record the operator semantics, source-local execution rules, replay contract, and multi-view composition used in the reported system.

\subsection{Typed Transition Semantics}
\label{app:operators}

Each state-bearing interaction declares three public components: a predicate key, a value type, and a temporal operator. The VLM predicts only the typed value consumed by that operator. Table~\ref{tab:operators} lists the concrete transition family instantiated in our experiments.

\begin{table}[ht]
\centering
\small
\setlength{\tabcolsep}{3pt}
\begin{tabular}{llll}
\toprule
Operator & Domain & Initial state & Transition\\
\midrule
\textsc{Add} & $\mathbb{N}_0$ & $0$ & $c\leftarrow c+\delta$\\
\textsc{Max} & $\mathbb{N}_0$ & $0$ & $c\leftarrow\max(c,z)$\\
\textsc{Persist} & Boolean & false & $e\leftarrow e\lor z$\\
\textsc{First} & timestamp & unset & $\tau\leftarrow\min(\tau,z)$\\
\textsc{Replace} & typed literal & unset & $s\leftarrow z$\\
\bottomrule
\end{tabular}
\caption{Concrete temporal operators used by \method{}. \textsc{Add} consumes a local increment, while \textsc{Max} consumes a cumulative estimate.}
\label{tab:operators}
\end{table}

The operators encode distinct temporal roles. \textsc{Add} accumulates evidence local to the current interval; \textsc{Max} retains a cumulative estimate; \textsc{Persist} records whether a condition has ever been established; \textsc{First} preserves its earliest supported timestamp; and \textsc{Replace} tracks the latest instantaneous value. The distinction between \textsc{Add} and \textsc{Max} is semantic rather than type-level: both consume nonnegative integers, but they assume different meanings for the predicted value.

\begin{figure*}[t]
\centering
\includegraphics[width=\textwidth]{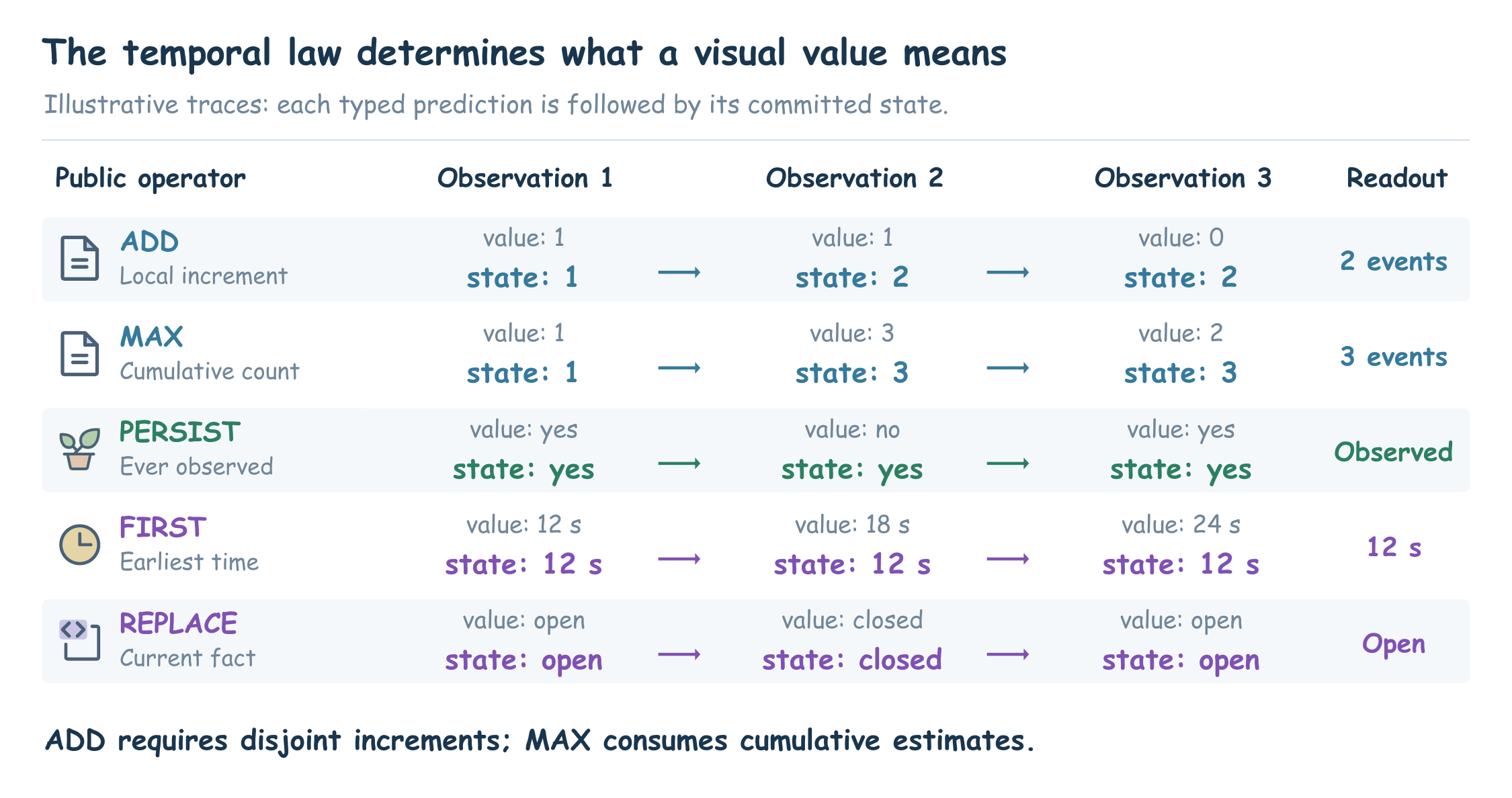}
\caption{Illustrative traces of the primitive state operators. Each column pairs a typed prediction with its committed value. \textsc{Add} consumes disjoint increments; \textsc{Max} consumes cumulative estimates. \textsc{Persist} retains positive evidence, \textsc{First} takes the earliest supported timestamp, and \textsc{Replace} follows the current value. Initial states are zero, false, or unset as specified in Table~\ref{tab:operators}. These are schematic values, not benchmark predictions. The additional deployed OST replay policy is described in Appendix~\ref{app:qualitative-replay}.}
\label{fig:appendix-state-laws}
\end{figure*}

Figure~\ref{fig:appendix-state-laws} illustrates these primitive operators over three successive updates.

Before commitment, the compiler checks that the predicate, value type, and operator are mutually compatible. A valid transaction modifies only the addressed register. An invalid transaction leaves the previously committed state unchanged and remains recorded as a failed update.

Validation enforces the declared transition semantics; visual accuracy is measured against reference answers.

\subsection{Source-Local Execution and Replay}
\label{app:state-replay}

State is isolated by visual source. Two independent streams therefore maintain disjoint namespaces even when their predicates have the same textual form. Within one source, predicates mapped to the same key reuse the corresponding register across timestamps.

For every state-bearing interaction, the runtime records the raw model output, the compiled transaction, its source-local key, validation status, and the resulting committed value. Accepted transactions can subsequently be replayed in their recorded temporal order without invoking the VLM again.

Within the transactional executor, each accepted transaction carries a stable identity. Repeated delivery of an already committed identity is idempotent and does not apply the same state update twice.

This replay contract enables the execution interventions used in our analysis. Resetting state removes cross-query persistence while retaining every generated value. Reversing the recorded transaction order changes temporal execution while preserving the same predictions. Namespace interventions change which source-local registers can interact, again without regenerating perceptual outputs.

\subsection{Frozen Multi-View Deployment}
\label{app:frozen-bundle}

The complete system uses one Qwen3-VL-8B language/state checkpoint together with two fixed visual surfaces: a recent-detail surface and an expanded-history surface. The two surfaces share all language and state parameters and differ only in their declared visual-merger configuration. They constitute two visual interfaces over one shared language/state model. Surface identity is fixed by the acquisition path rather than by benchmark task type, task group, or reference answer.

Each surface policy selects a proposal and executes it in its own source-local register before composition. The video path uses cumulative \textsc{Max} and Boolean \textsc{Persist}. The map $G$ selects a task-formatted readout and decodes it into the declared integer or Boolean value; the executor then commits once to the separate global register. Local \textsc{Add} increments are used in the counting diagnostic.

\stitle{Causal view construction.}
Recent-4 samples four frames from the latest four-second window. The event view reserves these frames and adds up to 16 historical frames; compact history reserves eight frames from the latest eight seconds and adds up to 24 historical frames. Historical candidates come from a 64-point uniform causal-prefix grid, excluding points within the reserved recent window. After combining and deduplicating the candidates, the selector uses flattened $16\times16$ RGB probes. Novelty is the cosine distance between consecutive probes; curvature is the cosine distance between successive probe displacements. Candidates receive the score $0.20R(n)+0.45R(c)+0.10R(t)$, where $R$ is average-tie rank normalization to $[0,1]$. Historical slots take the highest scores, breaking ties by later timestamp and then smaller frame index. Selected frames are returned chronologically. Compact-history frames have a maximum edge of 224 pixels; its recent frames retain their decoded resolution.

\stitle{Acquisition and proposal selection.}
The policy acquires the event view when the recent answer's first-token, full-vocabulary top-two probability margin is at most $0.9976896$. It acquires compact history when the raw recent and event answers disagree. Starting with the recent answer, it considers event and history proposals in order, replacing a different answer when its margin minus the currently retained margin exceeds $-0.5058786571025848$.

On OVO, $G$ compares the raw expanded-policy answer with the reference Recent-4 answer. If they differ, it selects the expanded policy's state-aware readout; otherwise it selects the alternate recent-detail readout. Invalid video values preserve state and retain the raw answer. A committed count without a matching answer option also retains the current raw answer.

The recent-detail surface uses the visual merger learned by the state-interface checkpoint, while the expanded-history surface retains the reference merger. No language or state tensor changes between surfaces. The exact acquisition policy and merger identities are frozen before evaluation.

\begin{figure*}[t]
\centering
\includegraphics[width=\textwidth]{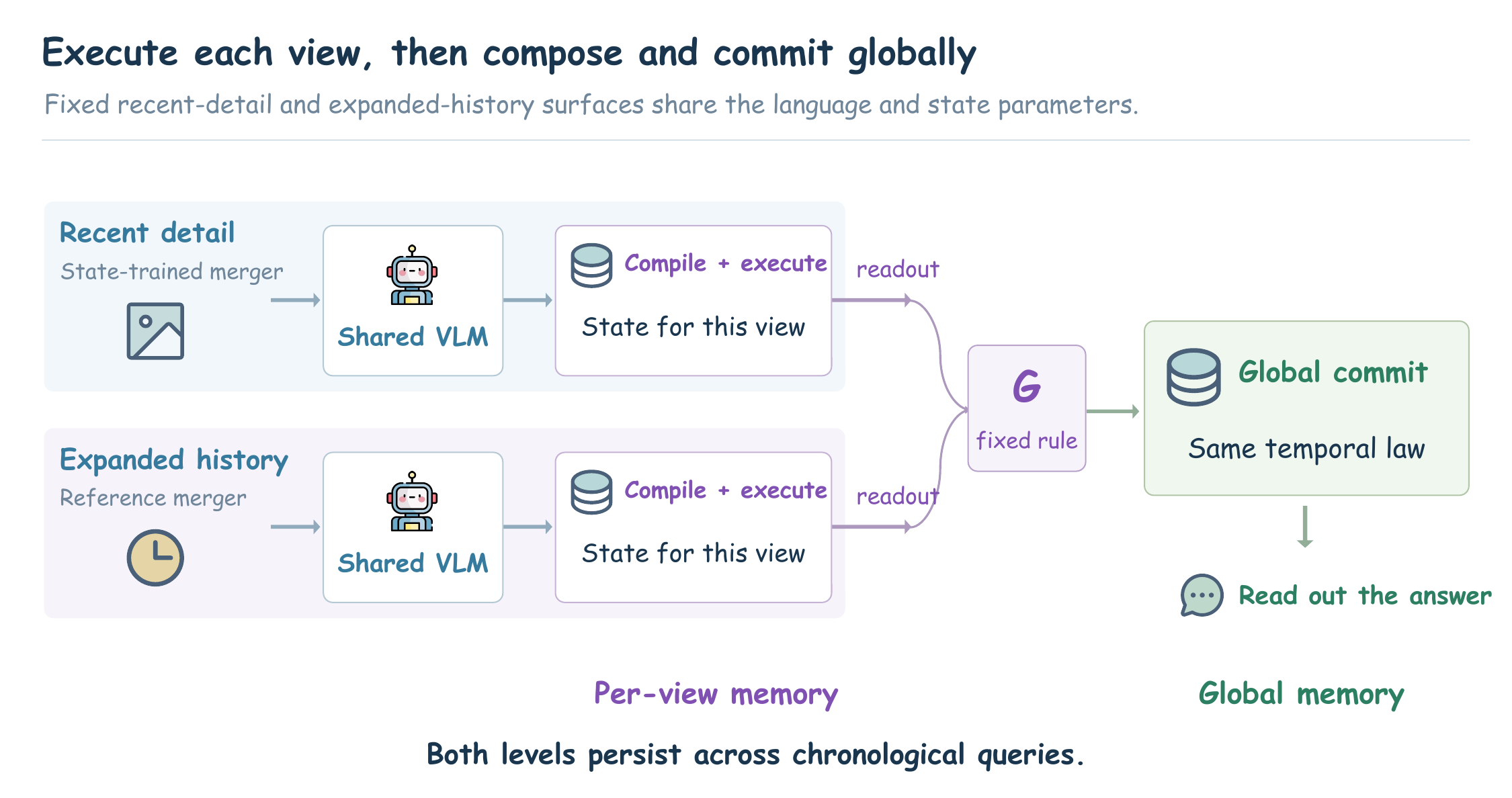}
\caption{Expanded view of state-aware composition. The two fixed visual surfaces share language and state parameters while using their declared visual mergers. Each proposal is compiled and executed in its own persistent register before its readout reaches the fixed composition map $G$. The composed proposal is committed to global state before answer readout. Both levels retain state across chronological queries.}
\label{fig:appendix-multiview}
\end{figure*}

Figure~\ref{fig:appendix-multiview} expands the per-surface and global execution paths.

\subsection{Answer-Space Consistency}
\label{app:answer-consistency}

For StreamingBench multiple-choice queries, a separate answer-space consistency policy takes a majority over the state-aware adaptive-policy answer, the state-aware binary-verification answer, and the option-orbit answer. A unique winner is selected; a tie retains the unpermuted Recent-4 answer. Option-orbit voting uses the unpermuted prediction and three nonidentity cyclic rotations, maps all predictions back to canonical labels, and applies the same unique-winner rule with the unpermuted prediction as fallback. Selection consumes predictions without task labels or gold answers.

The reported OST path applies entity reconciliation after native typed readout. Scene-local records retain existence, observed-by bounds, first-discovery times, and counts. An existing discovery record has priority; otherwise an observed-by bound can lower the proposed time. Counts respect the stored count and positive-existence floor. Auxiliary records update after readout, while primitive registers remain unchanged. Appendix~\ref{app:qualitative-replay} illustrates cross-query reuse and the resulting conflict priority.

\stitle{Per-query model calls.}
Across the complete evaluation populations, mean/maximum model calls are 2.43/12 on StreamingBench, 3.12/4 on OVO-Bench, and 1.00/1 per OST turn. Shared view predictions are counted once; verification and option rotations add calls, while deterministic compilation and replay add none.

\section{State-Oriented Supervision and Training}
\label{app:training}

This section summarizes how perceptual updates are supervised. Controlled target-formulation and runtime studies use separate checkpoints, as described below.

\subsection{Supervision Mixture}
\label{app:training-mixture}

Training combines ordinary question--answer records, annotated OST state targets, and local action-count targets. State examples supervise the typed values consumed by declared temporal operators; ordinary QA retains the direct-answer interface.

\subsection{Target Construction}
\label{app:data-provenance}

OST targets use annotated answers paired with the predicates and operators specified by public questions. Local counts use causal visual windows and completed action annotations. Training sources are separated from evaluation sources.

\subsection{Model Initialization}
\label{app:checkpoint-lineage}

The final system refines a shared Qwen3-VL-8B language/state model for typed visual updates. Its fixed recent and expanded visual surfaces use distinct visual mergers while sharing that model. The controlled checkpoints used below serve separate ablations.

\subsection{Matched Target Formulations}
\label{app:target-formulations}

The matched target-formulation study isolates the effect of the supervised output surface. It uses 9,848 OST records selected from 5,169 scenes in the 6,390-scene training pool, with zero known scene-ID overlap with the evaluation scenes. Each underlying example is rendered with one of three target formulations:

\begin{itemize}[leftmargin=*,itemsep=1pt,topsep=2pt]
    \item \textbf{Direct answer:} predict the final response to the current question.
    \item \textbf{Full transaction:} predict the predicate, temporal operator, and visual value as one serialized update.
    \item \textbf{Typed value:} predict only the value consumed by the compiler-declared transition.
\end{itemize}

The three paired runs share causal images, questions, sample weights, initialization, and optimization. Thus, the intervention changes the target formulation and its corresponding output/readout interface while holding the underlying visual training population fixed.

The average serialized targets contain 33.85 characters for direct answers, 66.44 for full transactions, and 17.34 for typed values. The typed formulation therefore tests whether the state-learning effect survives when the executable wrapper is removed from the learned output surface.

The typed formulation achieves comparable native performance to full transactions with a substantially shorter output, without requiring the model to generate executable syntax.

\subsection{Operator-Balanced Control}
\label{app:operator-balanced}

We train a separate checkpoint on the native/OST core with balanced state-operator supervision to support runtime and interface interventions.

This checkpoint is used only for controls that modify execution, history access, or interface structure. Its predictions and scores are kept separate from both the three-seed matched target-formulation study and the final deployment system.

\section{Experimental Protocol}
\label{app:experimental-protocol}

This section specifies the evaluation populations, causal inference policies, and checkpoint identities underlying our experiments. We distinguish complete-system performance from the controlled interventions used to attribute improvements to perceptual update learning and persistent state execution.

\subsection{Evaluation Populations and Scoring}

We evaluate \method{} on the complete locked populations of StreamingBench, OVO-Bench, and OST-Bench. Table~\ref{tab:evaluation-protocol} summarizes their evaluation units and official scoring rules.

\begin{table}[ht]
\centering
\small
\setlength{\tabcolsep}{3pt}
\begin{tabular}{lll}
\toprule
Benchmark & Population & Scoring\\
\midrule
StreamingBench & 2,495 queries & Query accuracy\\
OVO-Bench & 3,035 queries & Task--group macro\\
OST-Bench & 1,386 scenes & Official item score\\
\bottomrule
\end{tabular}
\caption{Locked benchmark populations and official evaluation metrics. OST contains 10,165 chronological turns and 9,528 officially scored items.}
\label{tab:evaluation-protocol}
\end{table}

All benchmarks use their official evaluators. StreamingBench scores the complete query population, while OVO-Bench applies its task-then-group aggregation. OST is processed chronologically across 10,165 turns, of which 9,528 contribute to the official aggregate.

Failed programs and unanswered queries remain in their original evaluation populations. Execution validity is tracked separately from official answer correctness, including scorer-specific empty-answer cases. This distinguishes successful state execution from successful task completion.

\subsection{Causal Access and Inference Policies}

At each query timestamp, the inference policy accesses only observations available within the causal prefix. Sampling, preprocessing, and answer interfaces follow the declared policy for each experiment.

In Tables~\ref{tab:sb-compare} and~\ref{tab:ovo-compare}, numerical frame budgets denote maxima per model call, not total frames across a query's calls. Entries in fps retain the reported sampling rate; a dash indicates an unreported or inapplicable budget.

Recent-frame, expanded-history, and full-history settings are distinct observation policies. Execution contrasts hold the checkpoint and visual evidence fixed; access and visual-grounding interventions vary the specified evidence. Mixed-interface controls use the declared 200,704-pixel inference processor budget, while the matched OST learning study retains its processor policy across target formulations.

The final deployment uses fixed visual surfaces and deterministic answer-space consistency (Appendices~\ref{app:frozen-bundle} and~\ref{app:answer-consistency}). Language and state parameters remain shared across surfaces; benchmark-specific prompts and answer interfaces follow their evaluation protocols.

\stitle{External comparisons.}
Published systems retain their reported populations, frame policies, and inference budgets and serve as contextual performance references. The fixed paper baselines are 80.52 on StreamingBench, 58.97 on OVO-Bench, and 50.52 for the matched OST reference.

The Gemini~3 Flash row in Table~\ref{tab:ovo-compare} is taken from Table~1 of \citet{li2026vam}, which reports the baseline under the OVO-Bench protocol without a model-specific frame budget.

Mage-VL, JoyAI-VL-Interaction, and VideoChat3 retain the complete rows reported in Tables~8--9 of \citet{zeng2026onestreamer}. That evaluation reports 2,498 StreamingBench real-time queries and selects each re-evaluated model's configuration by its whole-benchmark aggregate. Its selected frame budgets are not given per row.

\subsection{Controlled Comparisons and Checkpoint Identity}

Our experiments distinguish three configuration families, each serving a different purpose.

\stitle{Matched perceptual learning.}
The matched OST target-formulation study trains three paired seeds under matched visual data and optimization. Direct-answer, full-transaction, and typed-value formulations differ in their supervision targets and corresponding output/readout interfaces. The typed-value mean is $61.15\pm0.73$.

\stitle{Frozen-prediction execution.}
The OVO attribution study uses one frozen checkpoint and identical generated typed values across local and persistent readouts. Evidence, parameters, and model calls remain fixed. Persistent execution reaches 67.06, separating state composition from perceptual generation.

\stitle{Complete deployment system.}
The final system combines shared language/state parameters with two fixed visual-merger surfaces, state-aware composition, and answer-space consistency. The same frozen deployment bundle reaches 87.01 on StreamingBench, 70.09 on OVO-Bench, and 60.29 on OST-Bench.

\stitle{Scope of attribution.}
The matched OST study measures state-oriented learning together with its output/readout interface. The frozen OVO study isolates persistent execution on fixed predictions. Complete-system results measure the combined deployment policy, including visual access and answer-space consistency.

Contrasts are formed within their declared configuration family, evaluation population, and inference protocol. Source-held temporal-partition experiments use a separate diagnostic population (Appendix~\ref{app:composition}).

\section{Per-Task Benchmark Results}
\label{app:per-task}

Table~\ref{tab:ost-main} reports the complete per-subtype OST-Bench comparison summarized in Table~\ref{tab:summary} of the main text; the per-task OVO-Bench comparison is Table~\ref{tab:ovo-compare} in the main text. Published rows retain their released populations, frame policies, and inference budgets; the shaded rows are the frozen \method{} system and its Qwen3-VL-8B-Instruct reference on the complete locked populations.

\begin{table*}[!t]
\centering
\scriptsize
\setlength{\tabcolsep}{1.65pt}
\renewcommand{\arraystretch}{1.10}
\resizebox{\textwidth}{!}{%
\begin{tabular}{lcccccccccccccccc}
\toprule
& \multirow{3}{*}{Avg}
& \multicolumn{4}{c}{Agent State}
& \multicolumn{5}{c}{Agent Visible Information}
& \multicolumn{6}{c}{Agent--Object Spatial Relationship}\\
\cmidrule(lr){3-6}\cmidrule(lr){7-11}\cmidrule(lr){12-17}
& &
\multicolumn{2}{c}{Position}
& \multicolumn{2}{c}{Orientation}
& \multicolumn{2}{c}{Existence}
& \multicolumn{1}{c}{Quantity}
& \multicolumn{1}{c}{Diversity}
& \multicolumn{1}{c}{Order}
& \multicolumn{3}{c}{Direction}
& \multicolumn{3}{c}{Distance}\\
\cmidrule(lr){3-4}\cmidrule(lr){5-6}\cmidrule(lr){7-8}
\cmidrule(lr){9-9}\cmidrule(lr){10-10}\cmidrule(lr){11-11}
\cmidrule(lr){12-14}\cmidrule(lr){15-17}
Model &
& JUD. & EST. & JUD. & EST. & JUD. & TEMP. & CNT. & JUD. & JUD.
& JUD. & TEMP. & EST. & JUD. & TEMP. & EST.\\
\midrule
\rowcolor{GroupGray}
\multicolumn{17}{l}{\textit{Representative published MLLMs}}\\
Claude-3.5-Sonnet
& 47.8 & 65.1 & 36.2 & 50.6 & 30.3 & 85.8 & 67.0 & 57.9
& 57.1 & 60.0 & 39.2 & 18.3 & 21.9 & 43.8 & 54.9 & 19.0\\
Gemini-2.0-Flash
& 49.5 & 59.7 & 27.3 & 56.7 & 36.5 & 89.6 & 70.8 & 59.4
& 78.7 & 55.6 & 42.0 & 17.8 & 21.5 & 45.5 & 48.9 & 27.1\\
Gemini-2.0-Flash (Thinking)
& 54.2 & 57.4 & 36.3 & 61.1 & 33.4 & 88.1 & 74.8 & 61.9
& 63.6 & 73.1 & 50.9 & 51.7 & 23.3 & 51.1 & 56.8 & 22.7\\
GPT-4.1
& 53.4 & 64.2 & 30.7 & 60.8 & 33.2 & 90.8 & 78.0 & 60.6
& 82.1 & 70.8 & 51.5 & 23.4 & 28.6 & 44.9 & 53.6 & 23.9\\
GPT-4o
& 48.7 & 55.6 & 20.5 & 45.6 & 33.6 & 90.6 & 75.6 & 59.8
& 78.2 & 59.6 & 46.1 & 19.5 & 21.4 & 43.1 & 50.6 & 20.4\\
InternVL-2.5-38B
& 50.8 & 56.7 & 31.8 & 54.6 & 38.4 & 91.7 & 74.7 & 61.1
& 79.8 & 62.1 & 42.1 & 20.6 & 27.7 & 42.7 & 42.5 & 28.1\\
InternVL-2.5-78B
& 51.1 & 60.8 & 34.4 & 49.9 & 40.7 & 90.7 & 74.4 & 65.9
& 77.9 & 61.2 & 43.4 & 22.4 & 17.8 & 46.7 & 44.4 & 22.9\\
InternVL-2.5-8B
& 44.6 & 51.7 & 26.1 & 49.4 & 40.4 & 86.3 & 51.3 & 56.4
& 60.7 & 38.4 & 37.2 & 33.8 & 22.8 & 43.0 & 42.9 & 27.9\\
Qwen2.5-VL-72B
& 45.6 & 57.0 & 27.6 & 52.2 & 37.1 & 86.1 & 64.5 & 61.5
& 75.7 & 34.5 & 41.4 & 21.1 & 8.2 & 44.5 & 39.3 & 18.7\\
Qwen2.5-VL-32B
& 46.9 & 51.0 & 31.1 & 53.5 & 39.4 & 85.3 & 64.8 & 59.2
& 73.4 & 41.8 & 39.5 & 24.9 & 25.7 & 43.6 & 39.1 & 20.3\\
\midrule
\rowcolor{GroupGray}
\multicolumn{17}{l}{\textit{Our complete Qwen3-VL-8B evaluation}}\\
Qwen3-VL-8B-Instruct
& 50.52 & 58.65 & 42.12 & 55.32 & 37.02 & 83.29 & 64.07 & 61.37
& 76.33 & 59.71 & 43.35 & 25.57 & 21.58 & 48.27 & 44.93 & 34.47\\
\rowcolor{MethodBlue}
\textbf{\method{}}
& \textbf{60.29} & 63.10 & 48.75 & 61.64 & 43.82 & 93.69 & 77.18 & 77.78
& 83.15 & 71.82 & 47.29 & 33.91 & 44.07 & 54.47 & 58.91 & 32.95\\
\bottomrule
\end{tabular}}
\caption{\textbf{Fine-grained results on OST-Bench} (\%).
We follow the official Table~2 taxonomy of fifteen question subtypes.
Published rows use the released full split; the shaded rows report our complete
scene-disjoint evaluation. JUD., TEMP., CNT., and EST. denote judgement,
temporal localization, counting, and estimation.}
\label{tab:ost-main}
\end{table*}

\section{Controlled Experiments}
\label{app:controls}

This section provides the controls underlying the perception--dynamics factorization. We distinguish three interventions: changing the supervised output under matched training, changing the readout interface under a fixed checkpoint, and changing temporal execution while preserving generated predictions. Additional visual and runtime controls test grounding, state persistence, and execution validity. All comparisons follow the configuration identities specified in Appendix~\ref{app:experimental-protocol}.

\subsection{Matched State Supervision}
\label{app:learning-controls}

We first compare state-oriented supervision and readout under matched training conditions. The direct-answer, full-transaction, and typed-value runs share the same 9,848 OST training records, causal observations, initialization, optimization, and three paired seeds. Their supervised targets and corresponding output/readout interfaces differ.

\begin{table}[h]
\centering
\small
\setlength{\tabcolsep}{4pt}
\renewcommand{\arraystretch}{1.08}
\begin{tabular}{lrrr}
\toprule
Seed & Direct & Full update & Typed value\\
\midrule
59 & 52.07 & 60.53 & 60.79\\
60 & 54.24 & 60.39 & 60.67\\
61 & 54.09 & 61.68 & 61.99\\
\midrule
Mean & $53.46\pm1.21$ & $60.87\pm0.71$ & $61.15\pm0.73$\\
\bottomrule
\end{tabular}
\caption{Matched OST learning formulations on all 1,386 scenes and 9,528 official scored items. Each paired seed shares training data and optimization; means report sample standard deviations across three seeds.}
\label{tab:target_formulations}
\end{table}

Table~\ref{tab:target_formulations} reports the complete paired results. Direct answers reach $53.46\pm1.21$, full transactions reach $60.87\pm0.71$, and typed values reach $61.15\pm0.73$, with standard deviations across three seeds. Typed-value supervision improves over direct answering by 7.69 points on average, with positive differences in every seed.

\stitle{Unified typed-value readout.}
The seed-59 checkpoints are additionally evaluated through one shared value-only interface on the same 9,528 officially scored turns (Table~\ref{tab:shared-typed-value}). The evaluation uses identical causal image prefixes capped at 24 frames, a common typed-value prompt, greedy decoding, and no state execution or previous answers. Direct-answer, full-transaction, and typed-value training reach 51.83\%, 43.41\%, and 47.03\%, respectively. The typed model trails the direct model by 4.80 points, reversing the native ordering.

\begin{table}[t]
\centering
\footnotesize
\setlength{\tabcolsep}{4pt}
\renewcommand{\arraystretch}{1.08}
\begin{tabularx}{\columnwidth}{@{}>{\raggedright\arraybackslash}Xrr@{}}
\toprule
\textbf{Training target} & \makecell{\textbf{Native OST}\\\textbf{(\%)}} & \makecell{\textbf{Unified value}\\\textbf{(\%)}}\\
\midrule
Direct answer & $53.46\pm1.21$ & \textbf{51.83}\\
Full transaction & $60.87\pm0.71$ & 43.41\\
\tvsrow
\textbf{Typed value} & $\mathbf{61.15}\pm0.73$ & 47.03\\
\midrule
Typed $-$ direct (pp) & $\mathbf{+7.69}$ & $-4.80$\\
\bottomrule
\end{tabularx}
\caption{\textbf{State-oriented learning under native and unified readouts.} Native OST reports mean accuracy and sample standard deviation across three paired seeds. Unified value-only accuracy uses the seed-59 checkpoints, identical causal inputs, one common typed-value prompt, and no state execution. Bold marks the best result in each column.}
\label{tab:shared-typed-value}
\end{table}

\stitle{Output-length control.}
The serialized targets average 33.85 characters for direct answers, 66.44 for full transactions, and 17.34 for typed values. Typed values achieve the highest native mean with the shortest output surface and perform comparably to full transactions.

\stitle{Visual-grounding control.}
Removing visual inputs lowers typed-value accuracy by 15.13 points for seed 59 and 15.07 for seed 61, averaging 15.10 points. The questions and state interface remain fixed. For seed 59, typed supervision repairs 1,468 scored items and introduces 790 regressions; at the scene level, 710 improve and 316 decline.

Image removal establishes the visual grounding of the typed native interface. Across formulations, the reversed ordering under the unified readout shows that native effectiveness depends jointly on supervision and the readout interface.

\subsection{History and State Readout}
\label{app:history-controls}

We next examine whether exposing historical information through a generic memory interface provides the same benefit as executable state. Table~\ref{tab:matched_harness} compares direct dialogue, full-history memory, and typed execution within fixed checkpoints.

\begin{table}[t]
\centering
\small
\setlength{\tabcolsep}{4pt}
\begin{tabular}{lrr}
\toprule
OST interface & Base & Reference SFT\\
\midrule
Direct visual dialogue & 50.73 & 58.53\\
Full-history harness & 48.84 & 57.97\\
Typed persistent compiler & 27.04 & 60.26\\
\bottomrule
\end{tabular}
\caption{OST interface comparisons under fixed checkpoints and evaluation policies.}
\label{tab:matched_harness}
\end{table}

For the trained checkpoint, direct dialogue reaches 58.53, the generic full-history harness reaches 57.97, and typed persistent execution reaches 60.26. Historical access and executable state therefore yield different results under this controlled configuration.

The Base checkpoint exhibits a different interface sensitivity. Direct dialogue reaches 50.73, while imposing the typed compiler interface yields 27.04. Training on state-bearing targets is therefore important for obtaining useful values through the typed interface.

Historical evidence exposure and a trained state-oriented interface have distinct effects in this comparison, which evaluates each interface together with its readout.

\subsection{Execution on Fixed Predictions}
\label{app:execution-attribution}

The frozen OVO experiment isolates temporal execution after perceptual generation. The local and persistent arms reuse identical causal observations, model parameters, calls, and generated typed values.

Local readout reaches 64.74, while committing the same predictions through persistent state transitions reaches 67.06. A source-paired bootstrap that recomputes the official task-then-group macro gives a 95\% confidence interval of $[1.33,3.19]$ points for the 2.32-point gain. The difference is therefore attributable to executing the fixed values through persistent state in this configuration. The paired direct-answer interface scores 64.51, providing a separate reference for the typed output surface.

\subsection{Operator-Matched Answer Accumulation}
\label{app:operator-comparison}

We compare the two output interfaces on the complete OST population using one checkpoint and matched causal observations, processor budget, and decoding cap. Each arm retains its native prompt and history, so the comparison measures the complete interfaces under the same perceptual budget. The ordinary-answer arm parses its answer and receives the appropriate deterministic rule: maximum for cumulative counts, OR for persistent occurrence, minimum for first occurrence, and replacement for instantaneous values. The typed arm uses the compiler-declared value and the same corresponding law. Invalid values preserve prior state in both arms.

\begin{table}[t]
\centering
\footnotesize
\setlength{\tabcolsep}{3.5pt}
\renewcommand{\arraystretch}{1.08}
\begin{tabular*}{\columnwidth}{@{\extracolsep{\fill}}lrrr@{}}
\toprule
Operator & Items & Direct+law & Typed+law\\
\midrule
\textsc{Max} & 315 & 76.51 & 76.51\\
\textsc{Persist} & 1,299 & 93.92 & 94.30\\
\tvsrow
\textsc{First} & 1,021 & 75.51 & \textbf{81.10}\\
\textsc{Replace} & 6,893 & 50.48 & \textbf{51.54}\\
\bottomrule
\end{tabular*}
\caption{\textbf{Operator-matched OST comparison.} Both interfaces use the same checkpoint, causal observations, 200,704-pixel budget, decoding cap, and the appropriate declared law. Values are item accuracy (\%). \textsc{Add} is studied separately in the counting diagnostic.}
\label{tab:operator-matched-ost}
\end{table}

The complete official score moves from 58.53 to 58.60 when the operator-matched rules are applied to direct answers, and from 60.23 to 60.26 for typed values. Table~\ref{tab:operator-matched-ost} shows where the interface difference appears. \textsc{Max} is identical, and \textsc{Persist} differs by 0.38 points. Typed execution gains 5.58 points on \textsc{First} and 1.06 on \textsc{Replace}. The advantage therefore concentrates outside the simple monotone accumulators that a task-specific answer cache implements naturally. \textsc{Add} is absent from OST and is evaluated through local increments in the temporal-partition diagnostic (Appendix~\ref{app:composition}).

\stitle{Multi-view execution.}
The final system extends the same state semantics to multiple causal views. Table~\ref{tab:controlled-ablation}(c) in the main text reports the surface and execution-stage ablations; Table~\ref{tab:final-system-ablation} reports the answer-space consistency policy.

\begin{table}[h]
\centering
\small
\setlength{\tabcolsep}{4pt}
\renewcommand{\arraystretch}{1.10}
\begin{tabularx}{\linewidth}{@{}>{\raggedright\arraybackslash}Xr@{}}
\toprule
Frozen policy & SB\\
\midrule
Adaptive surface + state & 86.45\\
+ binary consistency & 86.65\\
+ policy consensus & 86.77\\
\rowcolor{AntPaleStrong}
\textbf{+ option orbit} & \textbf{87.01}\\
\bottomrule
\end{tabularx}
\par\vspace{3pt}
\footnotesize
Complete official population; no gold answers, question-type labels, or
task/group routing. This is an implementation analysis rather than evidence
for the perception--dynamics factorization.
\caption{\textbf{Final-system answer-space consistency on StreamingBench.}
Each row adds one task-blind consistency operation. The surface and
execution-stage ablations are Table~\ref{tab:controlled-ablation}(c).}
\label{tab:final-system-ablation}
\end{table}

Raw composition of the two visual surfaces reaches 68.45. Executing only at the global commit reaches 70.00, while executing only within individual views reaches 69.93. Applying both stages reaches 70.09. For the complete configuration versus raw composition, source-paired bootstrap resampling gives a 1.65-point gain with a 95\% confidence interval of $[0.76,2.31]$.

This gain uses the unrounded scores 70.09089697 and 68.44585293: their difference is 1.64504404 points, rounded to 1.65. Subtracting the displayed two-decimal scores instead gives 1.64 points.

Both execution-stage variants improve over raw composition, and the complete two-stage configuration achieves the highest score. Its additional gains over either single-stage variant are comparatively small. This ablation characterizes execution ordering within the final frozen system, separately from the fixed-prediction attribution checkpoint.

\subsection{Runtime Semantics}
\label{app:runtime-controls}

Persistent state introduces execution choices concerning temporal order, register identity, and transaction validity. We test these properties by replaying recorded predictions under controlled runtime interventions.

\stitle{Persistence and order.}
Using the separate operator-balanced OVO checkpoint, chronological execution reaches 67.62. Resetting state after each query reduces accuracy to 65.20, while reversing transaction order reduces it to 63.07. The generated perceptual values remain fixed across these interventions.

The results show that persistence and temporal ordering affect how recorded predictions contribute to later answers. They provide direct evidence that the execution policy is a consequential component of streaming state understanding.

\stitle{Source isolation.}
Merging otherwise independent source namespaces changes 231 predictions. This intervention preserves model outputs and modifies only the scope in which state variables interact. The resulting answer changes establish the operational importance of source-local state identity.

\stitle{Final-system runtime ablations.}
We additionally freeze the final system's proposals and intervene on the final state-commit stage across all three complete benchmark populations. Table~\ref{tab:controlled-ablation}(b) in the main text reports local readout, chronological execution, reset, reversal, and merged namespaces; all arms retain the complete official populations and require no new perceptual predictions. Earlier visual acquisition, per-view execution, and composition remain fixed. OST uses the deployed entity-level replay described in Appendix~\ref{app:qualitative-replay}. Its change from 60.2178 to 60.2874 measures the auxiliary reconciliation layer on top of native typed readout, affecting ten answers.

Resetting the final state returns the local score on all three benchmarks. Reversal has its largest effect on OVO, while merging unrelated source states reduces OST by 5.41 points. The additional final-commit gains are small because these proposals already include upstream state-aware inference; they complement the isolated state-execution contrast in Section~\ref{app:execution-attribution}.

\stitle{Program validity.}
On the frozen OVO attribution checkpoint, a separate interface control compares strict free-form Python generation with compiler-authoritative typed updates. Strict Python accepts 338 of 938 eligible OVO transactions and reaches 58.16, while the typed frontend accepts all 938.

This result characterizes the reliability of the executable interface. Compiler-owned syntax and validation allow typed visual values to be consumed under declared state laws without requiring the model to serialize executable program syntax.

\subsection{Interface and Visual Controls}
\label{app:interface-controls}

We further examine alternative state and memory interfaces using one operator-balanced OST checkpoint. Table~\ref{tab:interfaces} reports their complete evaluation results.

\begin{table}[t]
\centering
\small
\setlength{\tabcolsep}{4pt}
\begin{tabular}{lr}
\toprule
Operator-balanced interface & OST\\
\midrule
Direct visual dialogue & 57.53\\
Full-history harness & 57.21\\
State-only readout & 46.39\\
Strict executable program & 59.46\\
Typed persistent compiler & 59.45\\
\midrule
WorldIR, correct images & 49.64\\
WorldIR, images removed & 44.10\\
WorldIR, shuffled trajectory & 37.05\\
\bottomrule
\end{tabular}
\caption{OST interface and visual-grounding controls. WorldIR interventions share their own matched input interface.}
\label{tab:interfaces}
\end{table}

Direct dialogue and full-history memory reach 57.53 and 57.21, respectively. The state-only readout reaches 46.39, while strict executable programs and the typed compiler reach 59.46 and 59.45. The near-identical program and compiler scores show that their answer-level performance is comparable in this configuration.

\stitle{Visual interventions.}
Within the WorldIR interface, correct images yield 49.64. Removing images reduces accuracy to 44.10, and replacing the correct trajectory with a shuffled one reduces it to 37.05. These comparisons preserve the WorldIR readout while changing the visual evidence supplied to it.

All arms cover the complete 1,386-scene OST population. The recorded interfaces replay deterministically despite substantial differences in answer accuracy.

Together, these controls identify complementary benefits of matched state-oriented supervision and readout, and persistent execution under fixed perceptual predictions. Visual interventions measure grounding, while replay isolates how declared transitions maintain the supplied values.

\section{Temporal Compositionality}
\label{app:composition}

This section examines whether learned perceptual updates remain correct and compositionally consistent when the same visual process is partitioned at different temporal resolutions. We first define a source-held evaluation of counting updates, then compare independent and adjacent-family training. Additional full-population and full-mixture experiments assess the scope of the observed improvements.

\subsection{Evaluation Protocol and Metrics}
\label{app:composition-protocol}

For a counting process, let $\delta(I)$ denote the predicted number of completed events in interval $I$. For adjacent half-open intervals $I_L=(a,b]$ and $I_R=(b,c]$, correct local increments satisfy
\begin{equation}
    \delta((a,c])=\delta((a,b])+\delta((b,c]).
    \label{eq:appendix-count-composition}
\end{equation}
Events are counted by their annotated completion time, with each shared boundary owned by the preceding interval exactly once. The initial window additionally includes any event completed at frame zero.

We evaluate three frozen count pilots on a source-held population selected by a fixed hash. The original population contains 64 sources, disjoint from the 32-source development diagnostic. One source has a public media/annotation frame-clock mismatch and is recorded separately as a source-data error, leaving 63 evaluated sources without replacement.

For source $i$, let $c_{i,w}$ denote the final predicted count under temporal window width $w\in W=\{2,4,8\}$, and let $g_i$ denote its annotated count. We distinguish three evaluation criteria.

\stitle{All-partition correctness.}
A source is counted as correct when $c_{i,w}=g_i$ for every evaluated width. This measures whether the same source is correctly counted under all three temporal partitions.

\stitle{Pairwise agreement.}
We count equal final predictions across the three unordered width pairs. With $N=63$ sources, the denominator is $3N=189$. This measures agreement among final states induced by alternative partitions.

\stitle{Prediction range.}
We measure the variation of each source's predicted final count across widths:
\begin{equation}
    \mathcal{R}_{\mathrm{range}}=
    \frac{1}{N}
    \sum_{i=1}^{N}
    \left(
        \max_{w\in W} c_{i,w}
        -
        \min_{w\in W} c_{i,w}
    \right).
    \label{eq:partition-range}
\end{equation}
A smaller range indicates more stable final-count predictions across the evaluated partitions.

Each temporal window contains eight sampled causal frames. Changing its width also changes the number of windows, model calls, and total sampled observations. The measurements therefore characterize the complete visual-update process under the declared sampling policies. The 12-second test lies outside the diagnostic grid $W$ and is reported separately as a final-count transfer experiment.

\subsection{Partition Consistency and Transfer}
\label{app:composition-results}

Table~\ref{tab:temporal} in the main text compares independent multiscale supervision, adjacent-family training, and adjacent-family training with matched target distributions; all three pilots are frozen before source-held evaluation.

\stitle{Short-scale composition.}
Adjacent-family training increases pairwise agreement from 60/189 to 69/189. The unmatched adjacent pilot also reduces mean prediction range from 1.937 to 1.492, while retaining the independent pilot's 6/63 all-partition correctness.

Matching the target distribution increases all-partition correctness to 8/63 while preserving 69/189 pairwise agreements. Its mean range of 1.873 remains below the independent pilot's 1.937, although the unmatched adjacent pilot achieves the smallest range.

These results show that the training curriculum affects consistency and correctness differently. Adjacent examples improve agreement across partitions, while target-distribution matching yields a modest additional correctness gain.

\stitle{Larger-window transfer.}
The 12-second evaluation produces a different ordering. Independent multiscale training reaches 28/63 correct final counts, compared with 25/63 for unmatched adjacent training and 19/63 for target-matched adjacent training.

The reversed ordering separates temporal-scale transfer from short-scale agreement and correctness. As specified in Appendix~\ref{app:composition-protocol}, changing the window width also changes visual sampling and the number of update steps.

\stitle{Visual-grounding control.}
We replace the visual content with gray images while preserving each width's observation geometry. At 2, 4, and 8 seconds, all three program models and their corresponding fixed-executor readouts yield zero correct final counts.

The gray-image intervention establishes dependence on visual evidence within the tested counting interfaces.

\subsection{Adjacent-Family Construction}
\label{app:composition-construction}

The adjacent curriculum introduces explicit relationships among local temporal updates. For each selected family, we construct left, right, and merged intervals whose annotated increments obey Eq.~\ref{eq:appendix-count-composition}.

Families are assembled within existing count-program training blocks. A donor record can be replaced by a merged interval only when its original target equals $\delta_L+\delta_R$ and its source identity, counted action, frame count, and sample weight match the required record.

The target-matched construction preserves the marginal count-target distribution while changing the relationships among training observations. The independent and adjacent curricula can therefore be compared while controlling the frequency of the target values in the matched variant.

The construction changes the organization of supervision examples without introducing an additional consistency-loss term. The model continues to learn visual updates through the declared autoregressive training objective.

Adjacent-family supervision links local and merged observations through a common transition law; the evaluation measures its effect on final-count correctness and cross-partition stability.

\subsection{Broader Validation}
\label{app:composition-validation}

\stitle{Full-population temporal evaluation.}
We further evaluate all three frozen count pilots at 2-, 4-, and 8-second widths on the complete validation population. Of 4,593 sources, the same 123 public media-clock errors are retained separately at every width, leaving 4,470 evaluated sources without replacement.

Table~\ref{tab:count-full-validation} reports single-width correctness and cross-partition diagnostics for generated bounded updates and a fixed executor supplied with directly predicted increments. Both prediction interfaces receive the same sampled observations at each width.

\begin{table*}[t]
\centering
\small
\setlength{\tabcolsep}{4pt}
\begin{tabular}{llrrrrrr}
\toprule
Training curriculum & Readout & 2s correct & 4s correct & 8s correct & All correct & Pairwise equal & Mean range\\
\midrule
Independent & Generated & 1,212 & 1,501 & 1,779 & 490 & 4,817 & 1.651\\
Adjacent, unmatched & Generated & 1,292 & 1,668 & 1,722 & 575 & 5,698 & 1.300\\
Adjacent, matched & Generated & 1,341 & 1,779 & 1,779 & 534 & 4,950 & 1.644\\
\midrule
Independent & Fixed & 671 & 965 & 1,447 & 311 & 3,502 & 2.524\\
Adjacent, unmatched & Fixed & 1,258 & 1,735 & 1,968 & 579 & 4,828 & 1.775\\
Adjacent, matched & Fixed & 738 & 1,335 & 1,876 & 350 & 3,191 & 3.010\\
\bottomrule
\end{tabular}
\caption{Complete validation-population temporal ablations. Correctness columns have denominator 4,470; all-partition correctness requires the correct final count at all three widths. Pairwise agreement has denominator 13,410. Generated-update and fixed-executor readouts use distinct prediction interfaces.}
\label{tab:count-full-validation}
\end{table*}

Generated-update correctness increases from 33.58\% under independent training to 37.32\% with unmatched adjacent training and 39.80\% with target-matched adjacent training. The matched variant therefore improves generated-update correctness by 6.22 percentage points over independent training.

Across all three partitions, generated updates correctly count 490, 575, and 534 sources under independent, unmatched-adjacent, and target-matched-adjacent training. Pairwise agreement is 35.92\%, 42.49\%, and 36.91\%, respectively. Unmatched adjacency gives the strongest full-population cross-partition results, while target matching gives the highest four-second correctness. The ordering on the smaller 63-source diagnostic therefore does not extend uniformly to the complete population.

The fixed-executor results also depend on the curriculum: unmatched adjacency achieves 579 all-partition-correct sources, compared with 311 for independent and 350 for matched training. Its single-width correctness exceeds the generated-update interface at four and eight seconds. Visual-update accuracy, partition stability, and prediction interface must therefore be assessed separately.

\stitle{Full-mixture extension.}
We additionally test whether adjacent-family organization transfers to the complete state-learning mixture. The extension preserves all 26,273 training records, the relative order of non-count records, count-target frequencies, and total sample-loss mass.

The construction contains 1,622 selected families and 2,917 three-record count units. Each selected family is placed within one optimizer batch, with remaining positions filled from the original non-count sequence.

Under the fixed-input executable-state evaluation policy, the resulting checkpoint reaches 84.33 on StreamingBench, 66.02 on OVO-Bench, and 60.17 on OST-Bench. The corresponding reference state-learning configuration reaches 84.17, 67.06, and 60.26. The candidate therefore exhibits a small StreamingBench increase alongside lower OVO-Bench and OST-Bench scores.

These results establish the scope of the adjacent-family curriculum. It improves selected counting diagnostics and full-population four-second generated-update accuracy, while its extension to the complete training mixture yields mixed benchmark effects.

The diagnostics distinguish visual correctness, cross-partition stability, and temporal-scale transfer. Propositions~\ref{prop:partition} and~\ref{prop:consistency} formalize the composition requirement and its logical distinction from correct event counts.

\section{Theoretical Analysis}
\label{app:theory}

This section formalizes the perception--dynamics factorization underlying \method{}. Proposition~\ref{prop:entanglement} characterizes the ambiguity of answer-level supervision with respect to perceptual state updates. Proposition~\ref{prop:state_recursion} establishes conditions under which correct updates preserve a task-sufficient world state. Proposition~\ref{prop:partition} connects executable transition dynamics to temporal compositionality, while Proposition~\ref{prop:consistency} separates compositional consistency from perceptual correctness.

The analysis considers population objectives, idealized predictor classes, and declared primitive state-transition laws.

\subsection{Setting}
\label{app:theory-setting}

Let $\mathcal{H}$, $\mathcal{Q}$, and $\mathcal{Y}$ be the spaces of causal histories, supported queries, and answers. A fixed causal view constructor $g:\mathcal{H}\rightarrow\mathcal{V}$ exposes visual evidence $v=g(h)$. A deterministic reference map $a:\mathcal{Q}\times\mathcal{H}\rightarrow\mathcal{Y}$ supplies answers, and $z^\star:\mathcal{Q}\times\mathcal{H}\rightarrow\mathcal{Z}$ supplies the typed values required by declared transitions. A value may be a local increment, cumulative estimate, Boolean, timestamp, or typed literal.

For a finitely supported population $\mathcal{D}$, consider a representation $F(v)$ with an answer readout $G_A$ and a typed-update readout $J$. Both readouts return probability distributions. Define their population cross-entropy risks:
\begin{equation}
\begin{split}
    &\mathcal{R}_{\mathrm{ans}}(F,G_A)\\
    &\quad=\mathbb{E}_{(q,h)\sim\mathcal{D}}\big[
       -\log G_A(F(g(h)),q)[a(q,h)]\big].
\end{split}
\label{eq:theory-answer-risk}
\end{equation}
\begin{equation}
\begin{split}
    &\mathcal{R}_{\mathrm{upd}}(F,J)\\
    &\quad=\mathbb{E}_{(q,h)\sim\mathcal{D}}\big[
       -\log J(F(g(h)),q)[z^\star(q,h)]\big].
\end{split}
\label{eq:theory-update-risk}
\end{equation}
Here $G_A$ is distinct from the deployment composition map $G$. We use natural logarithms, allow idealized readouts to assign point masses, and omit loss components with zero weight. Zero-risk solutions are therefore attainable when the targets are consistent and the predictor class can represent them.

The distribution is \emph{answer-consistent} if equal views and queries imply equal answer targets. Update supervision is \emph{update-consistent} under the analogous condition for $z^\star$. The supervised answer support of a view is
\begin{equation}
\begin{split}
    \mathcal{A}_{\mathcal{D}}(v)=\big\{(q,a(q,h)):\;&
    (q,h)\in\operatorname{supp}(\mathcal{D}),\\
    &g(h)=v\big\}.
\end{split}
\label{eq:theory-answer-support}
\end{equation}
This fixed-view setting isolates the supervision question. A query-dependent acquisition policy can be studied by fixing the query and acquisition rule within the contrast.

\subsection{Transition Entanglement}
\label{app:theory-entanglement}

The first question is whether final-answer supervision necessarily requires a model to preserve distinctions needed by state evolution.

\begin{proposition}[Answer supervision can leave update distinctions unconstrained]
\label{prop:entanglement}
Let $h_1,h_2$ induce distinct views $v_1,v_2$, with $\mathcal{A}_{\mathcal{D}}(v_1)=\mathcal{A}_{\mathcal{D}}(v_2)$. Assume answer-consistent targets and an unrestricted representation/readout class on the finite view support.

\textnormal{(i)} There exists a zero-risk, answer-optimal solution whose representation merges the two views, $F(v_1)=F(v_2)$.

\textnormal{(ii)} Suppose $\mathcal{D}$ contains $(q_p,h_1)$ and $(q_p,h_2)$, each with mass at least $w>0$, and their typed targets are distinct. For $0\leq\lambda<1$, define
\begin{equation}
    \mathcal{R}_\lambda
    =\lambda\mathcal{R}_{\mathrm{ans}}+(1-\lambda)\mathcal{R}_{\mathrm{upd}}.
    \label{eq:theory-mixed-risk}
\end{equation}
Every solution with $F(v_1)=F(v_2)$ satisfies
\begin{equation}
    \mathcal{R}_\lambda\geq
    2(1-\lambda)w\log 2.
    \label{eq:theory-collapse-bound}
\end{equation}
If the update targets are also consistent, a joint zero-risk solution exists in this class. Thus every joint minimizer must distinguish $v_1$ and $v_2$.
\end{proposition}

\begin{proof}
For \textnormal{(i)}, an injective representation and point-mass readouts attain zero answer risk under answer consistency. Set $F'(v_2)=F(v_1)$ and leave other representations unchanged. Every supervised query--answer pair at $v_2$ also occurs at $v_1$. The unchanged readout therefore assigns probability one to its correct answer after the replacement.

For \textnormal{(ii)}, a merged representation gives $J$ the same representation and query at both points. Let $p$ be that output distribution and let $z_1\neq z_2$ be the two targets. Their loss contribution is at least
\[
    (1-\lambda)w\big(-\log p[z_1]-\log p[z_2]\big).
\]
Since $p[z_1]+p[z_2]\leq1$, their product is at most $1/4$, giving the bound. Under the final realizability assumptions, an injective representation and consistent point-mass readouts recover every target and achieve zero joint risk.
\end{proof}

\stitle{Interpretation.}
Proposition~\ref{prop:entanglement} formalizes an objective-level source of \emph{transition entanglement}. Answer supervision can admit representations that merge observations requiring distinct typed updates. Supervising the update value introduces an explicit separating constraint whenever those values differ. Different values imply different downstream states when the declared operator distinguishes them for an admissible prior state.

\stitle{Remark 1 (The view may hide the update).}
If two histories expose the same view and query but have different typed targets, they are identical inputs to the perceptual predictor defined here. Every predictor must return the same output distribution for both. With each example having mass at least $w$, their update risk is at least $2w\log 2$ by the same argument. Supervision cannot recover a distinction absent from the exposed inputs; evidence access and update learning therefore require separate controls.

\subsection{Executable Sufficient State}
\label{app:theory-state}

Let $\sigma:\mathcal{H}\rightarrow\mathcal{S}$ be a task-relative state map. We call it answer-sufficient when there exists a reference readout $\rho$ satisfying
\begin{equation}
    a(q,h)=\rho(\sigma(h),q)
    \label{eq:theory-task-sufficiency}
\end{equation}
for every supported history--query pair. For a stochastic reference answer, the corresponding definition requires its conditional answer distribution to depend on $h$ only through $\sigma(h)$.

Suppose extending $h_t$ with causal evidence gives $h_{t+1}$ and the reference state obeys
\begin{equation}
    \sigma(h_{t+1})
    =U_{o_t}\!\left(\sigma(h_t),z_t^\star\right).
    \label{eq:theory-reference-recursion}
\end{equation}
Here $U_{o_t}$ includes the addressed key and its declared transition law.

\begin{proposition}[Correct updates preserve sufficient state]
\label{prop:state_recursion}
Assume an answer-sufficient map $\sigma$, the reference recursion in Eq.~\ref{eq:theory-reference-recursion}, and $S_0=\sigma(h_0)$. If execution uses those declared keys and transition laws and receives $\hat z_t=z_t^\star$ for $t=0,\ldots,T-1$, then
\begin{equation}
    S_t=\sigma(h_t),\qquad t\leq T.
    \label{eq:theory-state-preservation}
\end{equation}
The reference readout $\rho(S_t,q)$ consequently recovers every supported answer at that time.
\end{proposition}

\begin{proof}
The assertion holds initially. If $S_t=\sigma(h_t)$ and $\hat z_t=z_t^\star$, then
\[
\begin{split}
    S_{t+1}
    &=U_{o_t}(S_t,\hat z_t)\\
    &=U_{o_t}(\sigma(h_t),z_t^\star)\\
    &=\sigma(h_{t+1}).
\end{split}
\]
Induction proves the state identity; Eq.~\ref{eq:theory-task-sufficiency} gives the answer identity. Under the stochastic definition, it gives equality of the reference conditional distributions instead.
\end{proof}

\stitle{Interpretation.}
The proposition characterizes executable state maintenance as a structured filtering process. The preceding state carries task-relevant history, perception supplies the new update, and execution propagates its consequences. Under correct initialization, predictions, and transition laws, the resulting state retains the reference information needed by the supported readout.

Monotone transitions can absorb individual updates: an already-true predicate remains true under another positive update, and a smaller cumulative estimate leaves a larger stored count unchanged. These behaviors motivate replaying fixed predictions under alternative execution policies.

\subsection{Typed Transition Algebra}
\label{app:theory-algebra}

For a fixed predicate and operator $o$, let $\oplus_o$ be an associative composition on legal update values. Here $S$ denotes that predicate's value state. We assume that its state action satisfies
\begin{equation}
\begin{split}
    U_o(U_o(S,z_1),z_2)
    =U_o(S,z_1\oplus_o z_2).
\end{split}
\label{eq:theory-transition-action}
\end{equation}
Commutativity is unnecessary because chronological order may matter.

The value laws in Table~\ref{tab:operators} satisfy this action with addition for increments, maximum for cumulative estimates, Boolean OR for persistent occurrence, minimum for first occurrence, and chronological right-projection for replacement. Unset first-occurrence state is treated as $+\infty$ for the minimum law. Right-projection returns the second update, so replacement is associative but order-sensitive.

These identities apply to primitive value transitions for one fixed predicate. Source namespaces and auxiliary entity reconciliation follow the separate execution rules in Appendix~\ref{app:state-interface}.

\begin{proposition}[Compatible updates yield partition invariance]
\label{prop:partition}
Let $I$ have a nonempty chronological partition into disjoint intervals $I_1,\ldots,I_n$. Suppose the predictions satisfy
\begin{equation}
    z(I)=z(I_1)\oplus_o\cdots\oplus_o z(I_n).
    \label{eq:theory-update-composition}
\end{equation}
If Eq.~\ref{eq:theory-transition-action} holds, sequential execution over that partition produces the same value state as executing the merged update once:
\begin{equation}
\begin{split}
    \operatorname{Fold}_o(S;z(I_1),\ldots,z(I_n))
    =U_o(S,z(I)),
\end{split}
\label{eq:theory-partition-invariance}
\end{equation}
where $\operatorname{Fold}_o$ applies $U_o$ successively in chronological order.
\end{proposition}

\begin{proof}
For one interval the claim is immediate. Applying Eq.~\ref{eq:theory-transition-action} merges the first two updates into their composed value. Repeating this reduction and using associativity merges the entire sequence into Eq.~\ref{eq:theory-update-composition}.
\end{proof}

\stitle{Implications for temporal evaluation.}
Proposition~\ref{prop:partition} gives a conditional algebraic guarantee. A learned perceptual model must additionally predict updates satisfying Eq.~\ref{eq:theory-update-composition}. For local event increments on adjacent half-open intervals, with each event assigned by its completion time, the required law becomes
\begin{equation}
    \delta((a,c])=\delta((a,b])+\delta((b,c]).
    \label{eq:theory-count-composition}
\end{equation}
This additive law applies to disjoint local increments, while cumulative estimates compose by maximum.

Equation~\ref{eq:theory-update-composition} specifies compatibility of local updates. The source-held experiments in Appendix~\ref{app:composition} measure final-count agreement across partitions, including the effects of visual sampling and learned predictions.

\subsection{Consistency and Perceptual Correctness}
\label{app:theory-consistency}

Temporal composition constrains relationships among predictions. Perceptual correctness additionally requires their agreement with the observed event process.

\begin{proposition}[Perfect compositionality can coexist with incorrect counts]
\label{prop:consistency}
Let $\delta^\star(I)$ be the true number of completed events in interval $I$, additive over adjacent disjoint intervals. For any integer $\alpha\geq0$, define
\begin{equation}
    \hat\delta_\alpha(I)=\alpha\,\delta^\star(I).
    \label{eq:theory-scaled-count}
\end{equation}
Then $\hat\delta_\alpha$ is additive over the same intervals. For every $\alpha\neq1$, it is incorrect on every interval with $\delta^\star(I)>0$.
\end{proposition}

\begin{proof}
Multiplication by $\alpha$ preserves addition:
\[
\begin{split}
    \hat\delta_\alpha(I_L\cup I_R)
    &=\alpha\big(\delta^\star(I_L)+\delta^\star(I_R)\big)\\
    &=\hat\delta_\alpha(I_L)+\hat\delta_\alpha(I_R).
\end{split}
\]
For a positive true count, multiplying by a nonnegative integer other than one changes the value.
\end{proof}

The all-zero predictor $\hat\delta_0$ is the simplest example: its final predictions agree under every partition and have zero range, yet every source with a positive event count is wrong.

\stitle{Implications for evaluation.}
Proposition~\ref{prop:consistency} motivates measuring final-count correctness, pairwise agreement, and prediction range separately. The source-held experiments examine how these complementary properties respond to adjacent-family training.

\subsection{Scope and Implications}
\label{app:theory-experiments}

These propositions characterize the objective, state-preservation, and composition properties of executable temporal updates. The controlled experiments examine how these properties manifest under learned visual predictions.

\section{Extended Related Work}
\label{app:related}

This section expands the positioning summarized in Section~\ref{sec:related}.

\stitle{World models and state estimation.}
World models learn latent or generative dynamics for prediction, planning, or control \citep{ha2018worldmodels,hafner2019planet,hafner2020dreamer,hafner2023dreamerv3,hafner2025mastering,bruce2024genie,sun2024latent}; \method{} instead estimates explicit world variables from observations under declared temporal semantics.

\stitle{Streaming video memory and evidence access.}
Streaming and long-video systems preserve history through sparse or hierarchical memories, compressed context, retrieval, recurrent bridges, and KV caches \citep{song2024moviechat,song2024moviechatplus,he2024malmm,zhang2024flashvstream,huang2025videochatonline,wang2025videollamb,liang2026oasis,xiao2026mukv,wen2026eventmemagent}, with SimpleStream \citep{shen2026simplestream} a strong recent-window reference and OneStreamer \citep{zeng2026onestreamer} reusing learned textual records. In medical VLMs, MedEyes progressively selects image regions \citep{zhu2026medeyes}, and MedSynapse-V builds latent diagnostic memory \citep{zhu2026medsynapse}. \method{} studies how values inferred from causal observations update persistent variables under declared temporal laws.

\stitle{Structured and executable state.}
Dialogue state tracking maintains explicit slot values \citep{wu2019trade,kim2020somdst,ouyang2020dstsc,heck2020trippy}, extended to visual objects by multimodal tracking \citep{le2022multimodaldst}; neuro-symbolic and programmatic visual reasoning composes perceptual predictions through modules, programs, and tools \citep{andreas2016nmn,yi2018nsvqa,gupta2023visprog,suris2023vipergpt,subramanian2023codevqa}. In language-only reasoning, SWAP verifies graph updates in planning \citep{xiong2025swap}, and MLR alternates structured planning with lower-level reasoning \citep{xiong2026mlr}. \method{} joins explicit state tracking with visual updates for causal streams: typed perceptual values update persistent variables under bounded executable laws.

\onecolumn
\section{Additional Performance Context}
\label{app:historical}

\begin{table}[ht]
\centering
\small
\setlength{\tabcolsep}{5pt}
\begin{tabularx}{\linewidth}{@{}>{\raggedright\arraybackslash}Xrrr@{}}
\toprule
Historical evaluated system & SB & OVO & OST\\
\midrule
Frozen \method{} system bundle & \textbf{87.01} & 70.09 & 60.29\\
State-interface SFT + adaptive access/state & 86.85 & \textbf{70.25} & 51.38\\
Joint SFT + adaptive access/state & 86.69 & 70.10 & 59.06\\
Spatial checkpoint A, direct dialogue & -- & -- & \textbf{64.01}\\
Spatial checkpoint A, typed compiler & -- & -- & 61.04\\
Spatial checkpoint B, direct dialogue & -- & -- & 63.55\\
Spatial checkpoint B, typed compiler & -- & -- & 62.13\\
\bottomrule
\end{tabularx}
\caption{Historical system scores. Bold entries are column maxima and need not belong to one system. Rows preserve their original checkpoint and inference-policy identity.}
\label{tab:historical}
\end{table}

Table~\ref{tab:historical} preserves earlier checkpoint and inference identities. The earlier state-interface system reaches 86.85 on StreamingBench and 70.25 on OVO, while its OST score is 51.38. Spatial checkpoint A reaches the highest recorded direct-dialogue OST score of 64.01, and spatial checkpoint B reaches the highest typed-compiler OST score of 62.13.

These rows describe the measured performance envelope across earlier systems. The main claims of \method{} use the matched target-formulation, fixed-prediction execution, and final-system experiments described above.

\section{Internal State-Execution Diagnostics}
\label{app:qualitative-replay}
% Non-floating horizontal figures: each sits between its case paragraphs.
\makeatletter
\newcommand{\inlinecasefigure}[3]{%
  \par\addvspace{3pt}\noindent\begin{minipage}{\linewidth}%
  \def\@captype{figure}\centering%
  \setlength{\abovecaptionskip}{4pt}%
  \includegraphics[width=0.95\linewidth]{figures/WorldCompiler-case-#1.pdf}%
  \caption{#2}\label{#3}%
  \end{minipage}\par\addvspace{3pt}}
\makeatother

Four recorded cases cover complete-system readout, cross-query reuse, monotone retention, and auxiliary reconciliation. Logged values are exact. The internal direct-answer and fact-memory controls use different models and visual-access policies; fixed-prediction execution attribution is reported in Appendix~\ref{app:controls}.

\subsection{Window Answers and an Already Composed Count}
In OVO soccer item 1613 (Figure~\ref{fig:case-main} in the main text), the direct control independently processes Recent-4 windows and returns $1,1,0$ at 20, 21, and 31 seconds. The end-card window receives zero. The \method{} composition instead supplies two at all three queries.

\inlinecasefigure{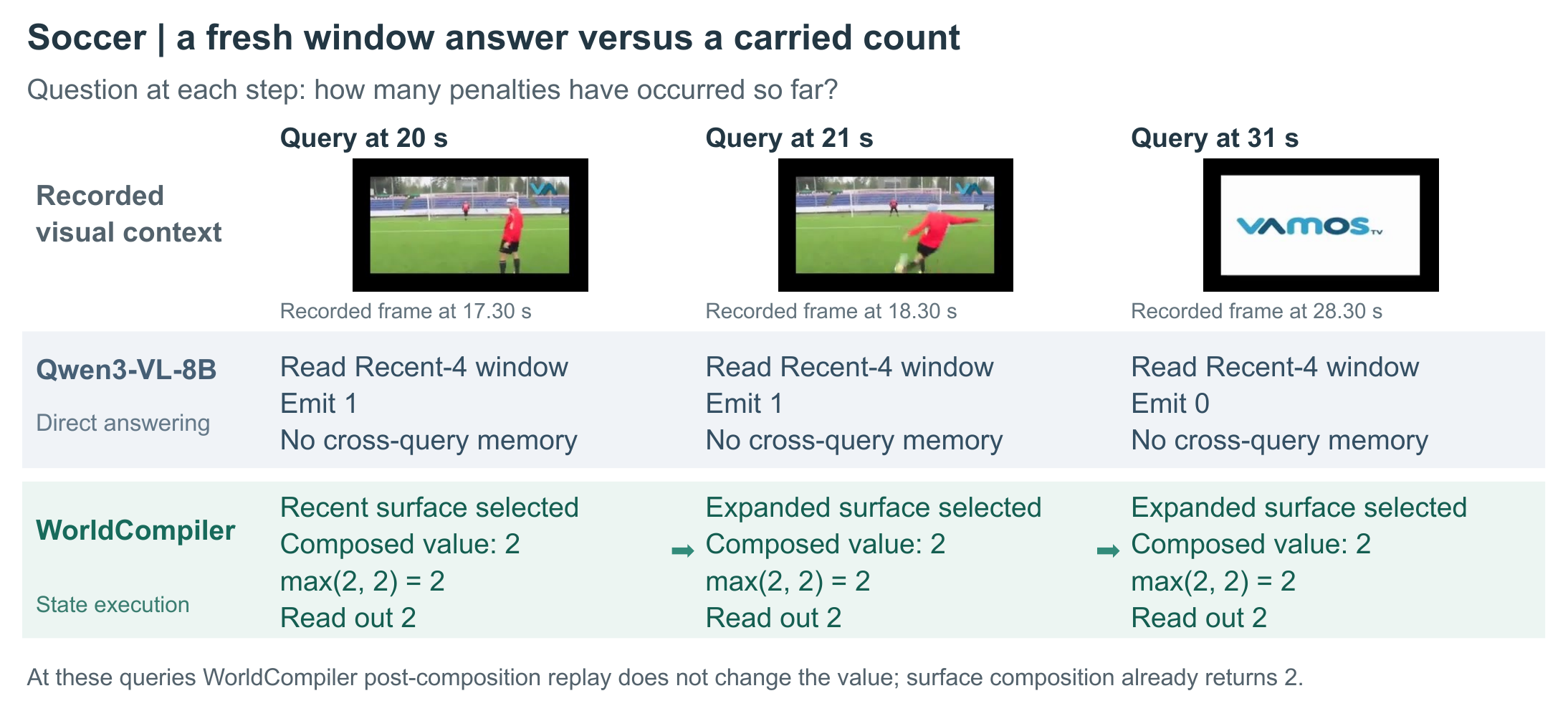}{Soccer: input-window answers, selected visual surfaces, composed values, and recorded maximum-register commits (full trace of the case summarized in Figure~\ref{fig:case-main}).}{fig:case-ovo-count}

The saved transitions are all $\max(2,2)=2$: the composed value already equals the final readout. The earlier 15-second answer is two against reference one; the three shown reference counts are two.

\subsection{Turning Co-Visibility into Reusable Entity Bounds}
For the bottle scene, the control returns three to the Turn-6 co-visibility question but its writer emits an empty fact list. At Turn~8 it again cites Turn~3 in its explanation and returns three.

\inlinecasefigure{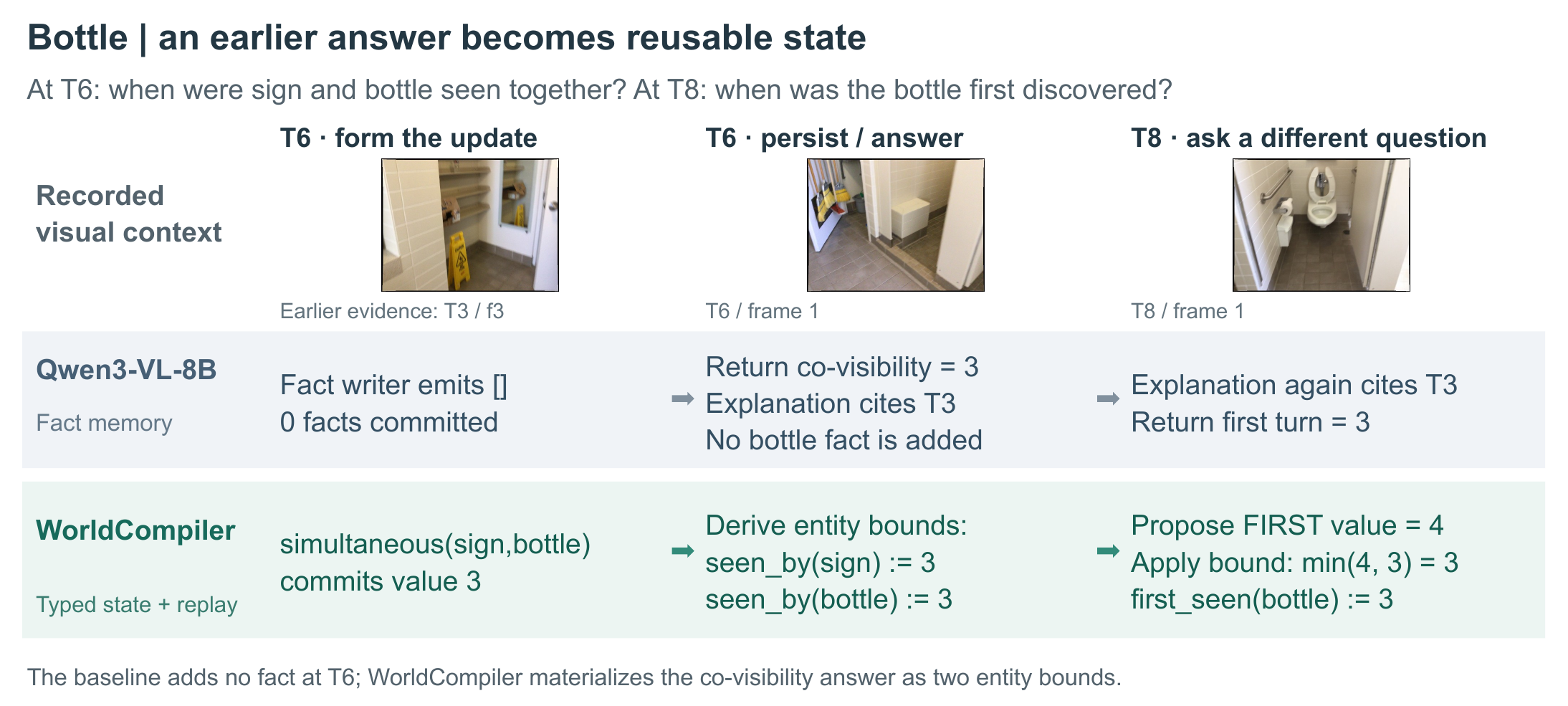}{Bottle: co-visibility establishes entity bounds reused by a later first-discovery query.}{fig:case-bottle}

WorldCompiler commits $\mathrm{simultaneous}(\mathrm{sign},\mathrm{bottle})=3$ and records bounds for both entities. The bottle bound converts the later first-discovery proposal from four to three, demonstrating cross-query reuse while both interfaces return the same answer.

\subsection{The Runtime Overrides a Later Weightlifting Estimate}
For OVO item 1608, the direct control returns one at 45, 60, and 75 seconds. The \method{} composed values are $2,3,2$, while its recorded state evolves $2,3,3$.

\inlinecasefigure{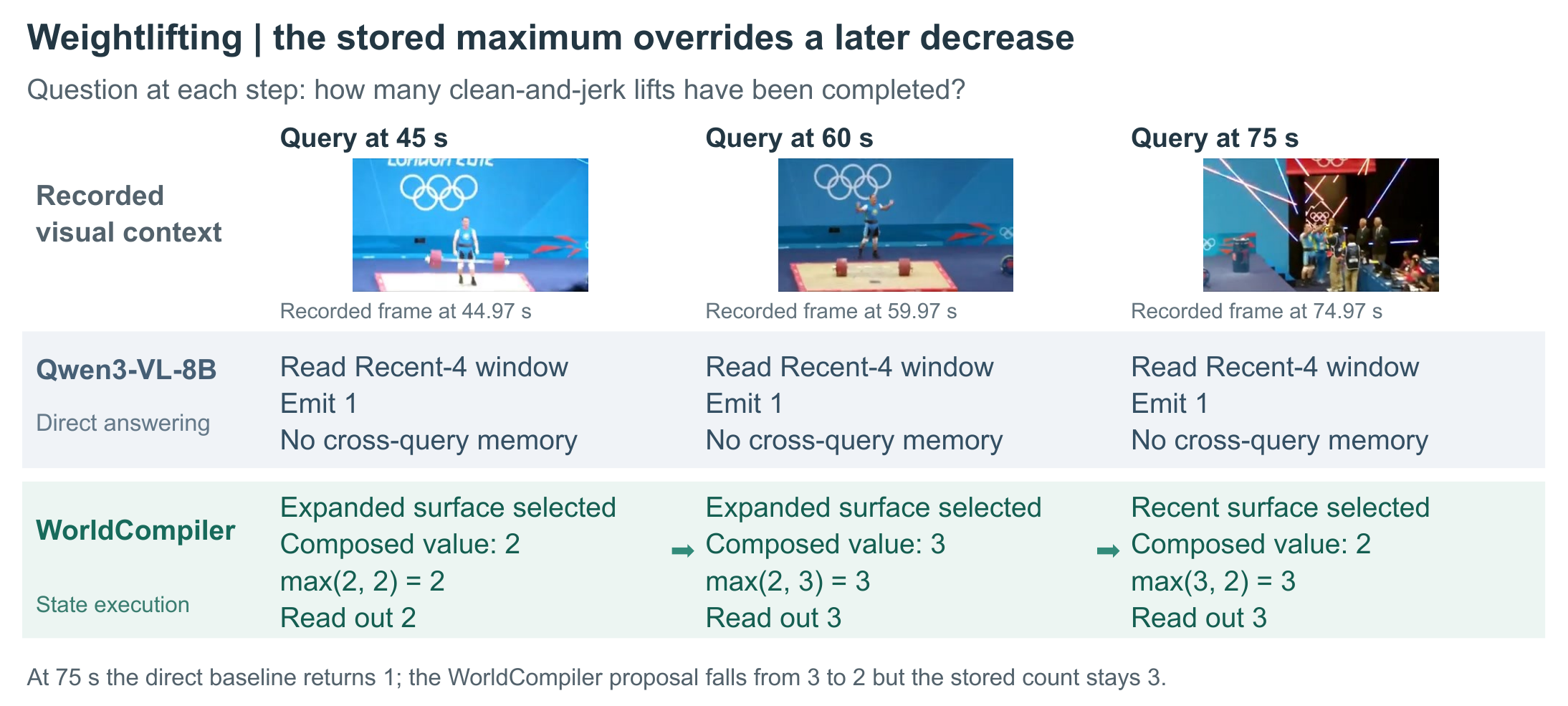}{Weightlifting: a later decrease is visible in the composed proposal but absent from the final readout.}{fig:case-ovo-overcount}

At 60 seconds, the expanded surface raises the stored count to three. At 75 seconds, the selected recent surface supplies two, but execution retains three. The reference is one throughout these queries. The figure separates the perceptual/composition error from the runtime's additional refusal to decrease.

\subsection{A Shelf Correction Is Generated and Then Suppressed}
Both methods first report shelf discovery at Turn~3. The control later adds a closer-to-shelf relation, then returns three again at Turn~5 with an explanation citing Turn~3.

\inlinecasefigure{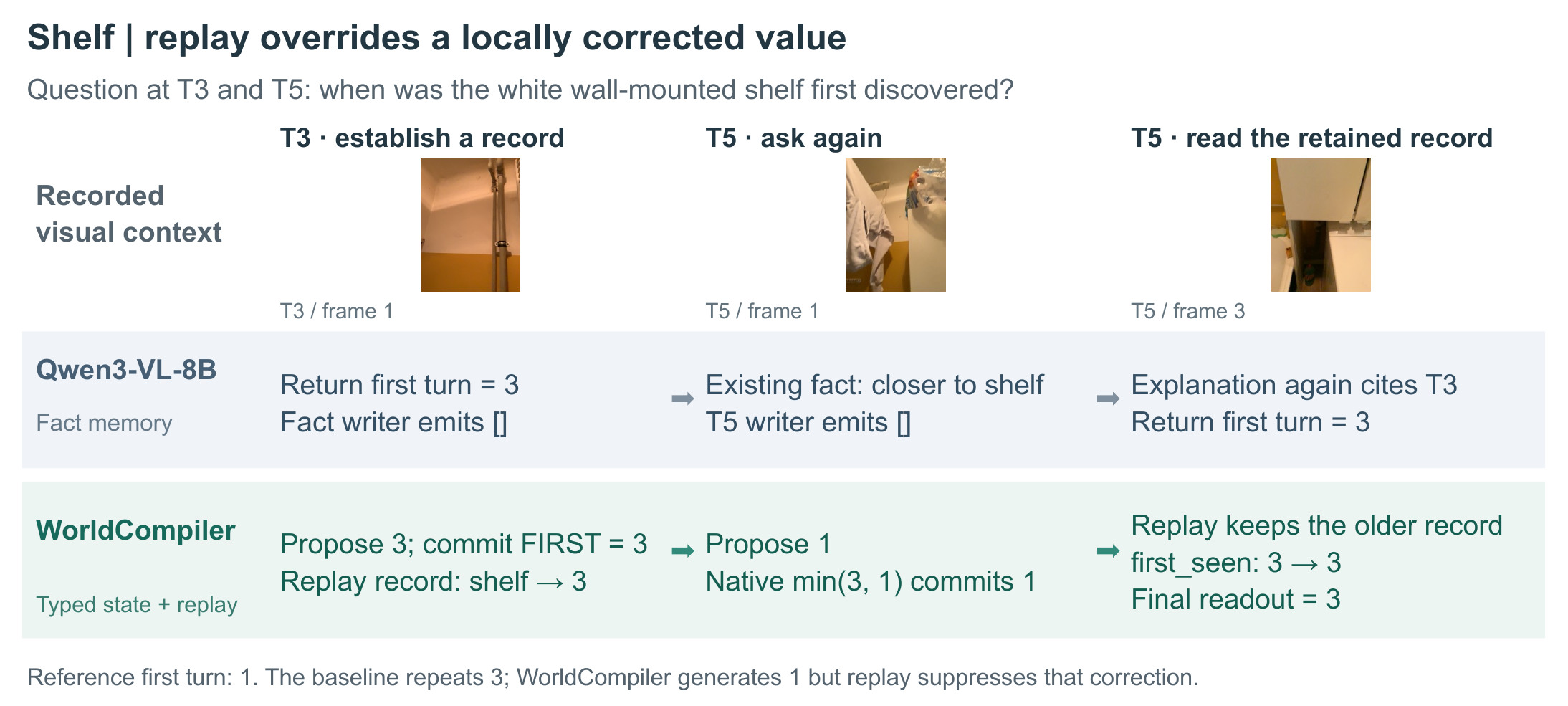}{Shelf: repeated control recall versus a locally corrected WorldCompiler value overridden by record retention.}{fig:case-shelf-failure}

At Turn~5, WorldCompiler generates one and its native minimum commits one, matching the reference. Replay nevertheless keeps the earlier record of three. This is a specific retention regression: the correct candidate exists in the trace but is not exposed as the final answer.

These examples expose the intermediate updates behind the answers. Complete-population execution effects and the ten-answer OST reconciliation intervention are reported in Appendix~\ref{app:runtime-controls}; the four cases are selected for mechanism coverage.

\end{document}